\documentclass{article}

\usepackage{arxiv}
\usepackage[authoryear,longnamesfirst]{natbib}
\usepackage{diagbox}
\usepackage{amsthm}
\usepackage{booktabs}  
\usepackage{longtable}
\usepackage{rotating}
\usepackage{multirow}
\usepackage{multicol}
\usepackage{graphicx}
\usepackage{subfig}
\usepackage{float}
\makeatletter  
\newif\if@restonecol  
\makeatother

\usepackage[linesnumbered,ruled,lined]{algorithm2e}
\usepackage{algpseudocode}

\SetKwFor{For}{for}{do}{endfor} 

\usepackage{CJKutf8}   
\usepackage[utf8]{inputenc} 
\usepackage[T1]{fontenc}    
\usepackage{hyperref}       
\usepackage{url}            
\usepackage{booktabs}       
\usepackage{nicefrac}       
\usepackage{microtype}      
\usepackage{lipsum}
\usepackage{graphicx}
\graphicspath{ {./images/} }

\title{Detecting fake news by enhanced text representation with multi-EDU-structure awareness}

\author{
 Yuhang Wang \\
  Data Science College\\Taiyuan University of Technology\\ Jinzhong, Shanxi, 030600, China\\
   \And
 Li Wang$*$ \\
  Data Science College\\Taiyuan University of Technology\\Jinzhong, Shanxi, 030600, China \\
  \texttt{wangli@tyut.edu.cn} \\
  \And
 Yanjie Yang \\
  Data Science College\\Taiyuan University of Technology\\Jinzhong, Shanxi, 030600, China\\
  \And
 Yilin Zhang \\
  College of Software\\ Taiyuan University of Technology\\Jinzhong, Shanxi, 030600, China\\
}

\begin{document}
\maketitle
\begin{abstract}
Since fake news poses a serious threat to society and individuals, numerous studies have been brought by considering text, propagation and user profiles. Due to the data collection problem, these methods based on propagation and user profiles are less applicable in the early stages. A good alternative method is to detect news based on text as soon as they are released, and a lot of text-based methods were proposed, which usually utilized words, sentences or paragraphs as basic units. But, word is a too fine-grained unit to express coherent information well, sentence or paragraph is too coarse to show specific information. Which granularity is better and how to utilize it to enhance text representation for fake news detection are two key problems. In this paper, we introduce Elementary Discourse Unit (EDU) whose granularity is between word and sentence, and propose a multi-EDU-structure awareness model to improve text representation for fake news detection, namely EDU4FD. For the multi-EDU-structure awareness, we build the sequence-based EDU representations and the graph-based EDU representations. The former is gotten by modeling the coherence between consecutive EDUs with TextCNN that reflect the semantic coherence. For the latter, we first extract rhetorical relations to build the EDU dependency graph, which can show the global narrative logic and help deliver the main idea truthfully. Then a Relation Graph Attention Network (RGAT) is set to get the graph-based EDU representation. Finally, the two EDU representations are incorporated as the enhanced text representation for fake news detection, using a gated recursive unit combined with a global attention mechanism. Experiments on four cross-source fake news datasets show that our model outperforms the state-of-the-art text-based methods. 
\end{abstract}


\section{Introduction}
It is a worldwide trend that more and more people get ample online news when they are surfing the web. However, with convenience, online platforms also provide a wide transmission range for fake news, causing catastrophic losses to individual life and society. For instance, during the outbreak of coronavirus disease 2019 (COVID-19), unprecedented amounts of fake news appeared on social media. According to reports, Facebook removed seven million posts for false coronavirus information, including content that promoted fake preventative measures \footnote{https://www.reuters.com/article/us-facebook-content-idUSKCN25727M}. The massive fake news created distrust among people, and hampered epidemic prevention and control measures. Thus, how to identify fake news efficiently has become a crucial problem. Several works have been proposed to tackle this problem\citep{LiWang_14}. They generally leveraged external information associated with news articles, such as comments and retweets \citep{shu2019defend,YANG2022116071,20212910664304}, time series \citep{10.1145/2806416.2806607}, user profile \citep{shu2018understanding,Lianbiao_3540} and so forth. Despite their success, the above approaches are inefficient in the early stage due to the labor-intensive data collection process. By contrast, text-based fake news detection is a convenient method that purely needs text content as input. 

In this study, we are concerned with text-based fake news classification task, which is conducive to fake news early detection. We formulate our task as a surpervised text classification problem, and train a classifier to map the input news text to its corresponding label to predict whether the news is fake or real. Previous text-based approaches typically learned various features in text, including manually designed linguistic styles and latent embeddings. The former first extracted shallow features (e.g. POS tags, Ngrams) by cumbersome feature engineering, then used machine learning methods, such as SVM and Logistic Regression, to identify fake news \citep{horne2017just}. Due to the low efficiency of feature engineering, some studies utilized deep neural networks to avoid manually designed features by automatically generating latent representations, thus improving the detection efficiency \citep{volkova-etal-2017-separating,wang-2017-liar,8864171}. Above approaches always focused on learning representations at word-level or sentence-level. In general, word-level models utilized isolated words to express news semantic, which will lead to inaccurate or unidiomatic expression, and cannot capture the text's meaning exactly because of the lack of context and coherence. One improvement is to use fixed-size window fetching for context information. \cite{zhang-etal-2020-every} applied fixed-size sliding window to words sequence to obtain co-occurrence relationships. The limitation is that it is too compulsory and mechanized to automatically find the optimal window size. An alternative option is to extend the window size to sentence length, but sentences are always coarse-grained and complex, lacking specific and detailed semantics expression. Some invalid noise in a long sentence may overwhelm the key information. 

Moreover, most text-based methods ignored the important role of text structural information in fake news detection task. Incorporating structural feature has been shown conducive to reveal the authenticity of text. \cite{vaibhav-etal-2019-sentence} explored text structure and discovered that it could affect the performance of fake news detection. They confirmed that there are factual jumps across sections in real news, i.e. sentences are highly cohesive if they belong to the same section, whereas fake news does not have this pattern. \cite{WANG2021114090} noted that local sequential order between consecutive sentences have a certain logic, switching the order would result in different meanings. However, they dealt with text structures in a simple way, such as constructing sentences into a fully connected graph. It may introduce noise to the semantic expression and reduce the detection effect. Therefore, how to further capture and use structural information to improve text representation still needs to be explored.

Enlightened by above discussions, our two main research questions are as follows:
\begin{itemize}
	 \item{} \textbf{RQ1}  Is there a better unit than word and sentence to express text semantic with high-quality? 
	 \item{} \textbf{RQ2}  How to utilize structural information to enhance the text representation for classification?
\end{itemize}

For \textbf{RQ1}, we introduce the Elementary Discourse Unit (EDU) as the basic unit of the text. It denotes the fine-grained subordinate clause, and is the intermediate granularity between word and sentence. Compared with word, it considers coherent semantics and expresses complete information. Compared with sentence, EDU is shorter and contains more specific information. Thus, we think EDU is a better unit than word and sentence.

\begin{figure*}[htbp]
	\centering
	\subfloat[Word-level text embeddings]{
		\includegraphics[width=0.46\linewidth,height=18em]{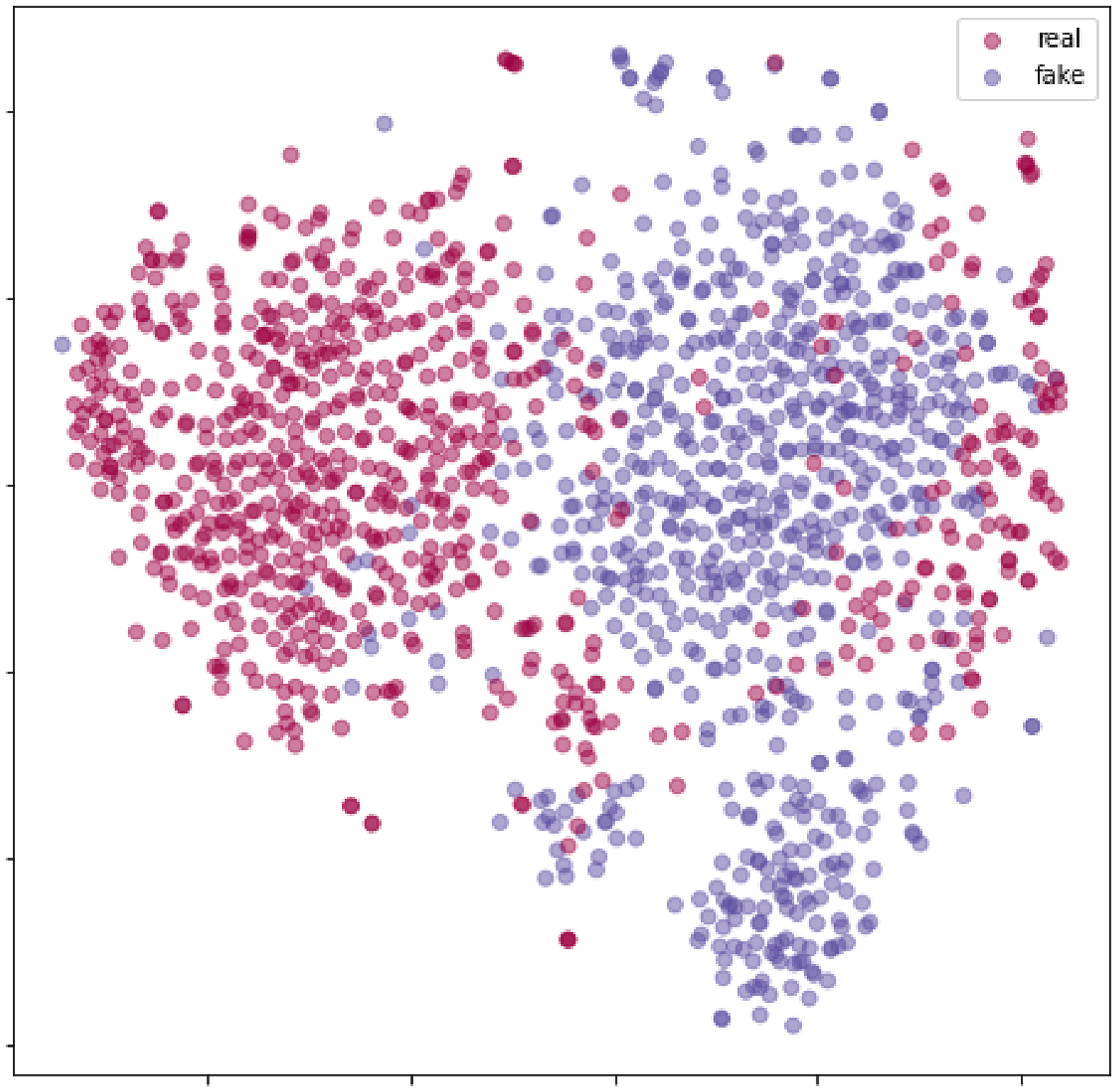}
		\label{fig:Word}
	}
	\quad
	\subfloat[Sentence-level text embeddings]{
		\includegraphics[width=0.46\linewidth,height=18em]{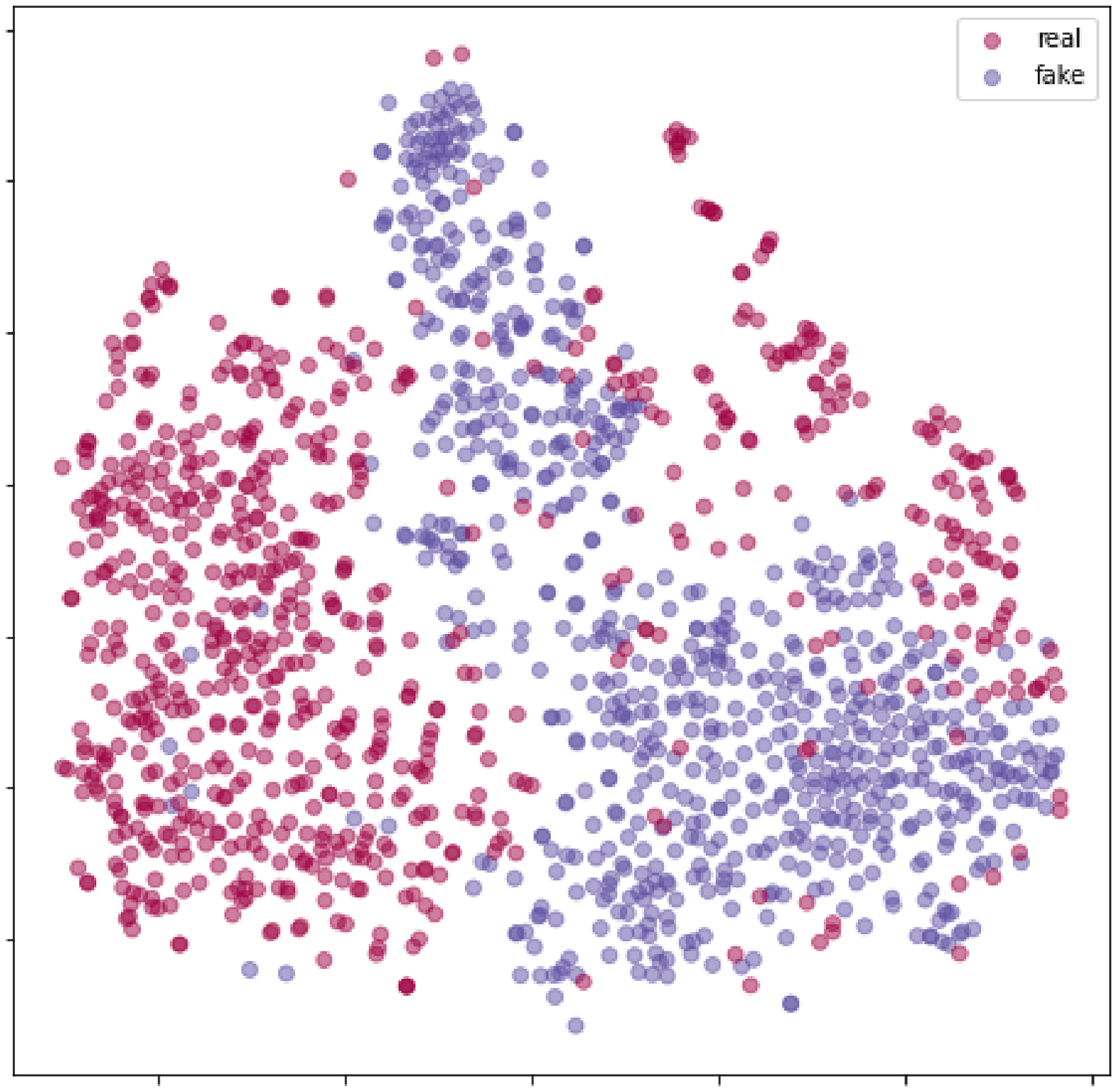}
		\label{fig:Sentence}
	}
	
	\subfloat[EDU-level text embeddings]{
		\includegraphics[width=0.46\linewidth,height=18em]{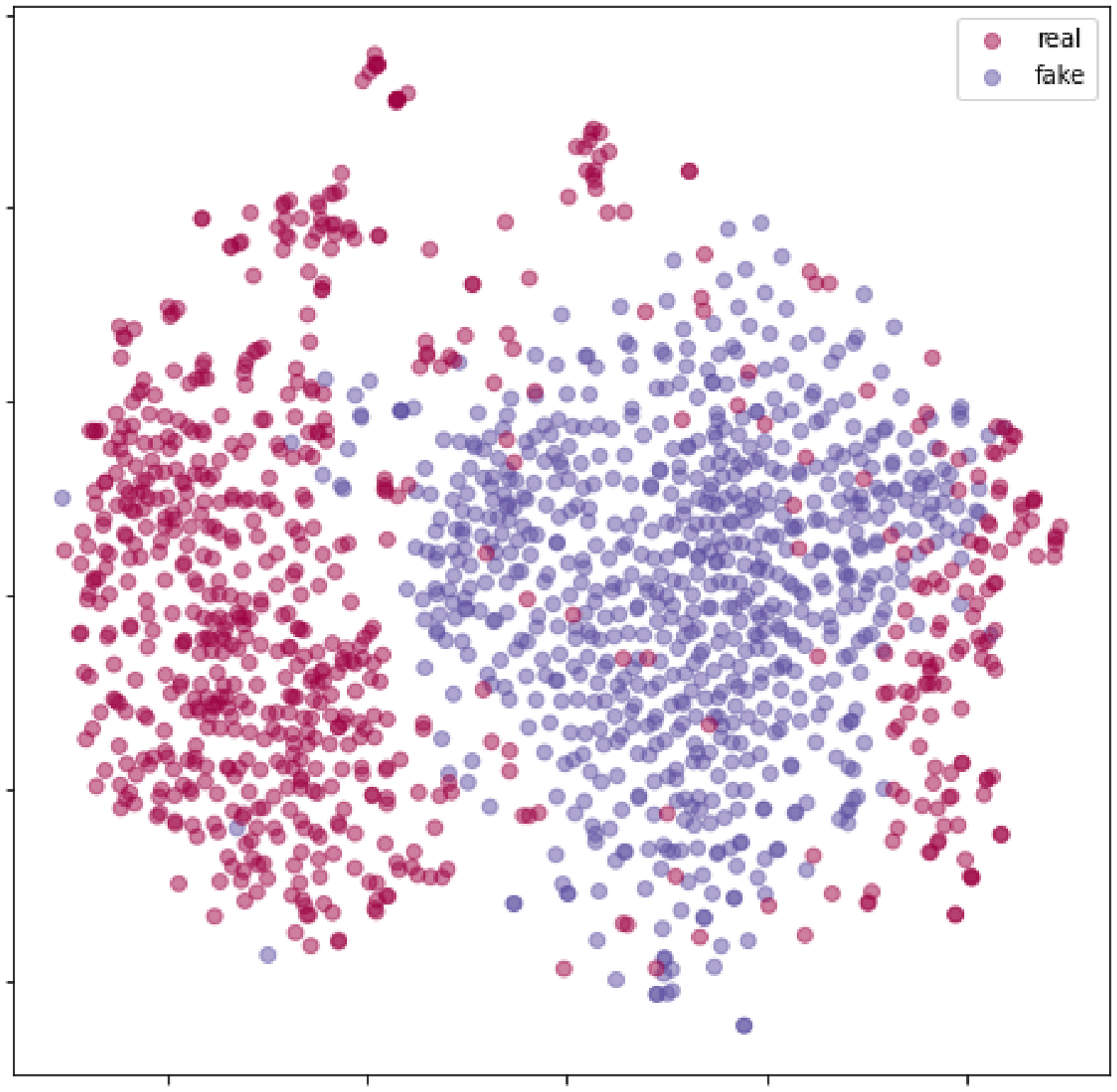}
		\label{fig:EDU}
	}
	\caption{t-SNE visualization of text embeddings, achieved by embedding words, sentences, and EDUs based on BERT \citep{devlin-etal-2019-bert}, followed by a max-pooling layer.}
	\label{fig:wse}
\end{figure*}

To reveal the impact of different granularities on fake news detection task, we conducted a visualization experiment on a well-known fake news dataset LUN-test \footnote{LUN-test is obtained from \citep{rashkin-etal-2017-truth}.}. Specifically, for each text, we first utilized BERT \citep{devlin-etal-2019-bert} to vectorize its different granularities of text units, including word-level, sentence-level and EDU-level. Then we obtained the text embeddings based on these three different granularities after a max-pooling layer and visualized them using t-SNE \citep{maaten2008visualizing}. As shown in Figure \ref{fig:wse}, each dot represents a news text, real news and fake news are shown in red and blue, respectively. Through data observation, we find that: (1) Text embeddings based on word-level units are cohesive in the same category, but the boundary is not clear and many dots are misjudged. (2) Sentence-level text embeddings are dispersed and poorly clustered. (3)
Obviously, EDU-level text embeddings provide a better cohesive effect and a clearer boundary than other granularities for this corpus. The results suggest that EDU contributes to high-quality representation learning compared with word and sentence.

For \textbf{RQ2}, based on the Rhetorical Structure Theory (RST) \citep{mann1988rhetorical}, there are multi-types of rhetorical relations (e.g., Contrast and Elaboration) between EDUs. These functional relations could describe the hierarchical discourse structure of the text and may reveal the underlying authenticity \citep{10.1002/asi.23216}. In this paper, we explore the EDU structures from two views. (1) The EDU sequential structure is constructed by arranging EDUs in the writing order. It can reflect the local coherence among consecutive EDUs, and some logic are implied in it, such as causal or contrastive relationship. (2) The EDU dependency graph structure is established, in which EDUs are connected by dependency rhetorical relations. The graph structure goes beyond mere sequential relationships and describes the global discourse dependencies between EDUs \citep{li-etal-2014-text}. It can express the global narrative logic of the text and help deliver the main idea truthfully.

\begin{figure*}[h]
	\centering
	\includegraphics[width=0.7\linewidth, height=5.8in]{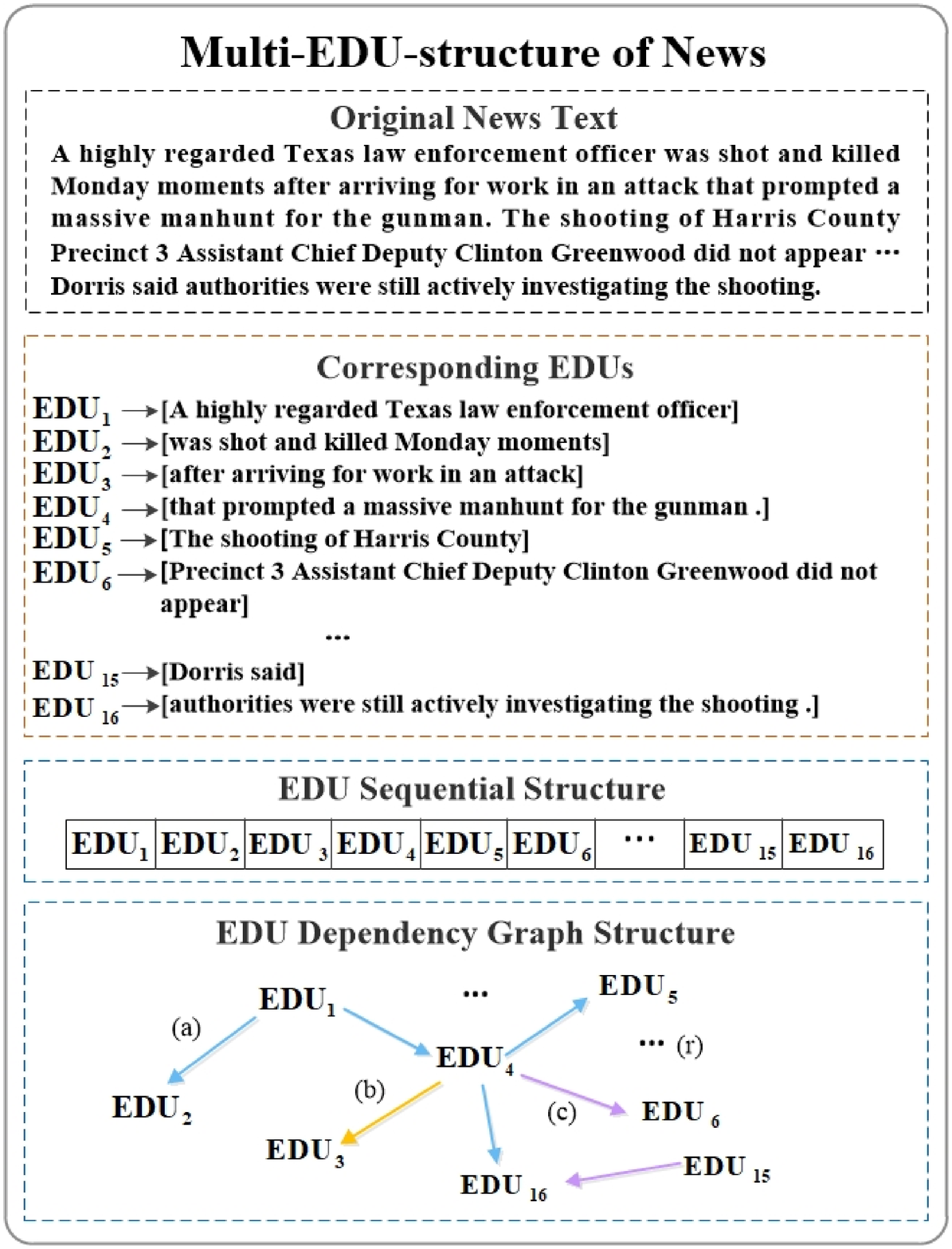}
	\caption{An example of EDU structure of news. (a) denotes the rhetorical relation named "Elaboration", (b) is the "Topic-comment" and (c) is the "Attribution".}
	\label{fig:example}
\end{figure*}

For instance, Figure \ref{fig:example} illustrates a news text from website\footnote{https://www.politifact.com/} and shows the multi-EDU-structures which we have built, including EDU sequential structure and dependency graph structure. The original news text are segmented into 16 EDUs. In the dependency graph structure, numerous rhetorical relations (Elaboration, Topic-comment and Attribution) are edges between EDUs, expressing the high-level organizational relationships in the text. This example shows that the text content expressed by two EDU structures is completely different. 

Based on the above discussions, we propose a multi-EDU-structure awareness model named EDU4FD that can effectively model text structural information from the perspective of EDU for fake news detection. EDU4FD captures the sequential-based EDU representations and the graph-based EDU representations simultaneously. It could consider both context information and discourse relationship between each neighboring EDUs, even if they located far off in the text.

The primary contributions of the paper include:

\begin{itemize}
	\item{} We introduce EDU, an intermediate granularity between words and sentences, which could help capture the fine-grained semantics of the whole news. And we propose a novel model EDU4FD for early fake news detection base on EDU.
	\item{} From the views of sequence and graph structure, we build the sequential-based EDU representations and the graph-based EDU representations. The former is gotten by modeling the coherence between consecutive EDUs with TextCNN. For the latter, we first build the EDU dependency graph, which describes the global discourse dependencies of text. Then a Relation Graph Attention Network (RGAT) is set to get the graph-based EDU representation. 
	\item{} We introduce the Gated Recursive Unit combined with the Global Attention mechanism (GRU-GA) network, which first enhances the EDU representation in top-down global reading order, then focuses on key EDUs to form a text representation for final prediction.
	\item{} Extensive experiments on four cross-source fake news datasets demonstrate that our approach is superior over the state-of-the-art methods.
\end{itemize}

\section{Related Work}
Text-based methods can identify fake news directly without auxiliary information, which is conducive to the fake news early detection. They generally focused on exploiting linguistic features and structural features. 

\subsection{Linguistic-based methods}
	Linguistic-based methods often extracted various features from words or sentences level and used machine learning \citep{horne2017just, perezrosas2018automatic} or deep learning models \citep{wang-2017-liar, volkova-etal-2017-separating, ijcai2017-545, 8864171} to capture linguistic knowledge and classify fake news. \cite{perezrosas2018automatic} extracted a set of manual features (e.g. Ngrams, Punctuation and Psycholinguistic features.) to train a linear SVM model. \cite{wang-2017-liar} developed a deep learning-based method to detect fake news using CNN and BiLSTM. \cite{GOLDANI2021102418} detected fake news using CNN with margin loss and severals word embedding models. \cite{8864171} were the first to use the BERT model to calculate sentence representation for fake news detection. The quality of features extracted by the above methods largely depends on the quality of the dataset, and the potential semantic information cannot be fully explored. Models were struggling to generalize to new text styles which are not available in training.  

\subsection{Structure-based methods}
Previous fake news detection methods mostly ignored the structural feature in the way of news text representation. Text structures could reflect potential pattern of fake news, which are not easy to be discovered and confronted by news forgers \citep{10.1002/asi.23216}. For structure-based methods, \cite{zhou2020fake} captured the writing style of fake news from Lexicon, syntax, semantics, and discourse level. At the discourse level, they used the rhetorical constituency tree to study the frequencies of relationships among sentences and utilized this style feature to detect fake news by machine learning methods (SVM, NB, LR, etc.). They acquired artificially designed feature extraction at the cost of efficiency and could not extract higher-level feature from the whole text perspective. Recently, graph-based methods demonstrated their promising performance on NLP tasks \citep{yao2019graph, zhang-etal-2020-every}. They modeled text as graph-structured data, and applied the Graph Convolutional Network (GCN) \citep{DBLP:conf/iclr/KipfW17} to achieve excellent text classification results via neighborhood propagation. \cite{vaibhav-etal-2019-sentence} first applied GCN to detect fake news solely based on its text content. They modeled the entire news text as a complete graph with sentences that are fully connected, and used a GCN to learn semantic information among pair-wise sentences. Based on them, \cite{WANG2021114090} proposed SemSeq4FD model, which fully considers the role of sentence relationships in enhancing text representation, including the global semantic relationship among far-off sentences, the local context sequential order between consecutive sentences, and the global sequential order of sentences in the whole text. The aforementioned works used sentence as the basic unit when learning text structural feature, and had shown competitive performance. However, long sentences may affect the expressive ability of the model as they are coarse-grained and lack specific information. To enrich the semantic expression, we introduce the Elementary Discourse Unit (EDU) into fake news detection task. EDU is the text unit segmented from sentence (See Figure \ref{fig:example}). It contains richer semantic information than pure word or sentence. Through fine-grained EDUs as well as the functional relationships between them, the structure of news text can be effectively described. In this paper, we focus on exploring the structural information implied in news text from EDU perspective and exploit both sequential structure and graph structure to enhance the text representation. 

\section{Problem Formulation}
\textbf{Task Definition}  The formal definition of fake news detection task is as follows: Given a news corpus $\mathcal{D}={\left\{ \mathcal{D}_i\right\}}^{N}_{i=1}$\, containing $N$ articles, we set $\mathcal{Y}={\left\{ {y}_i\right\}}^{N}_{i=1}$\, as a collection of corresponding labels indicating whether these articles are real or fake. Our goal can be described as a function $ f: \mathcal{D}\longrightarrow \mathcal{Y}$. It means that we seek to train a classification model $f$, mapping each article $\mathcal{D}_i$ to a label $\mathcal{Y}_i$ to distinguish whether the article is fake news or not.

\textbf{Major Notations} $\mathcal{D}_i={\left\{ {EDU}_{j}^{i}\right\}}^{|U|}_{j=1}$ is a news article from corpus $\mathcal{D}$, which has $|U|$ Elementary Discourse Units (EDUs), where each unit composed of ${T_j}$ words. We consider the discourse structure of $\mathcal{D}_i$ as an individual graph $\mathcal{G}_i = \left( \mathcal{V}, \mathcal{E} \right)$\,. $\mathcal{V}$ denotes the set of $|U|$ nodes, each of which is an $EDU$. Nodes are connected by specific relations. We use 19 kinds of rhetorical relations mentioned in \cite{li-etal-2014-text} to describe the intricate discourse structure contained in news. Based on these relations, $\mathcal{E}=\left\{\left({EDU}_{u},r,{EDU}_v\right) \mid {EDU}\in \mathcal{V}, r \in\mathcal{R}\right\}$\, is a set of edges between EDUs, where $\mathcal{R}$ is the rhetorical relation set. The key notations used in this paper are summarized in Table \ref{tab:notations}.
\begin{table*}[h!]
	\caption{The details of main notations in this paper}
	\label{tab:notations}
	\begin{tabular}{cl}
		\toprule
		\bf Notations&\bf Descriptions\\
		\midrule
		$\mathcal{D}={\left\{ \mathcal{D}_i\right\}}^{N}_{i=1}$ & A corpus $\mathcal{D}$ contains $N$ news articles \\
		${EDU_j^i}=\{\textbf{w}^{i}_1, \textbf{w}^{i}_2, \ldots, \textbf{w}^{i}_{T_j}\}$ &The $j$th $EDU$ in the article $\mathcal{D}_i$\,, composed of ${T_j}$ words \\
		$\mathcal{G}_i = \left( \mathcal{V}, \mathcal{E} \right)$ & Dependency discourse graph of article $\mathcal{D}_i$\,, $\mathcal{V}$ is node set, $\mathcal{E}$ is edge set \\ 
		$\mathbf{X}_C$, $\mathbf{X}_G$ & Sequence-based representation matrix and Graph-based representation matrix\\
		$\mathbf{z}$ &The text representation\\
		$\mathcal{Y}={\left\{ y_i\right\}}^{N}_{i=1}$ & The label set corresponding to $N$ news articles\\
		\bottomrule
	\end{tabular}
\end{table*}
\begin{figure*}[h]
	\centering
	\includegraphics[width=1\textwidth,height=3.2in]{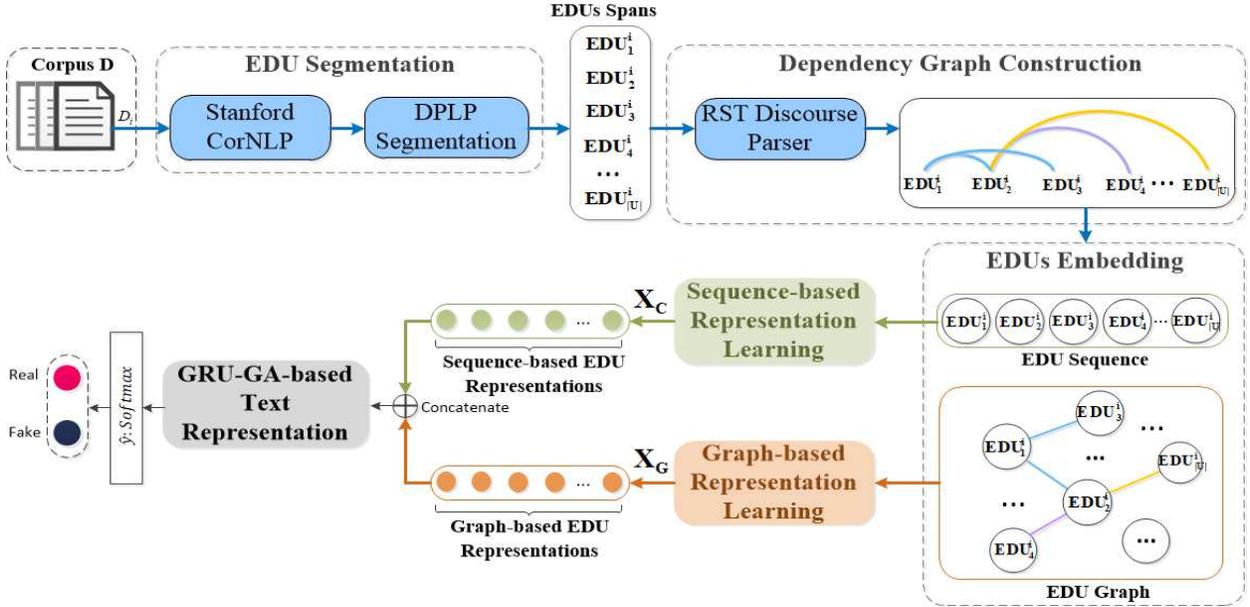}
	\caption{Overall framework of EDU4FD.}
	\label{fig:framework}
\end{figure*}

\section{Methodology}
In this section, we show how to implement the EDU4FD for detecting fake news based on EDU structure. The overall framework is illustrated in Figure \ref{fig:framework}, which consists of six modules namely EDU Segmentation and Dependency Graph Construction, EDUs Embedding, Sequence-based Representation Learning, Graph-based Representation Learning, GRU-GA-based Text Representation, and Inductive Classification. 

\subsection{EDU Segmentation and Dependency Graph Construction} \label{Section41}
In order to inject structural information to EDU4FD, this section describes how to segment EDU and construct the dependency discourse graph for a given article with the following two stages:
\begin{itemize}
	\item{\textbf{EDU segmentation}}: As shown in Figure \ref{fig:framework}, for an input text, we first adopt Stanford CorNLP\footnote{1. https://stanfordnlp.github.io/CoreNLP/} to split the text and tokenize words and sentences. Then, to accurately segment EDUs, we follow the pre-training model DPLP (Discourse Parsing from Linear Projection) proposed by \cite{ji2014representation}, which uses the gold EDU segmentation method to produce EDUs.
	\item{\textbf{Dependency Graph Construction}}: To obtain fine-grained dependencies between EDUs, we feed EDUs into the RST parser proposed by \cite{li-etal-2014-text} and parse them into a dependency tree. The tree consists of EDUs as nodes that are linked by rhetorical relations with functional meaning, such as Condition, Cause, and Explanation. By removing the root node that has no real semantic content, we convert it to a dependency graph structure, on which we can directly analyze the rhetoric relations among text units without considering the ambiguity in complex hierarchical constituency tree. 
\end{itemize}

\subsection{EDUs Embedding}  \label{Section4.2}
EDU4FD learns the raw EDU vectors via a bidirectional Gated Recurrent Unit (BiGRU) encoder. The BiGRU encoder contains $\overrightarrow{GRU}$ and $\overleftarrow{GRU}$, which could capture contextual information from forward and backward order. For a text $\mathcal{D}_i$ with $|U|$ EDUs, each ${EDU}_j^i$ consists of $T_j$ words, which can be mapped to the corresponding word vectors, i.e., ${EDU_j^i}=\{\textbf{w}_1^{j}, \textbf{w}_2^{j}, \ldots, \textbf{w}_{T_j}^{j}\}$. The representation of ${EDU}_j^i$ is calculated as follows:
\begin{equation}
\overrightarrow{\textbf{h}^{j}_{t}}=\overrightarrow{GRU}(\overrightarrow{\textbf{h}^{j}_{t-1}},\textbf{w}^{j}_{t})
\end{equation}
\begin{equation}
\overleftarrow{\textbf{h}^{j}_{t}}=\overleftarrow{GRU}(\overleftarrow{\textbf{h}^{j}_{t+1}},\textbf{w}^{j}_{t})
\end{equation}
\begin{equation}
 \textbf{h}^{j}_{t}=\overrightarrow{\textbf{h}^{j}_{t}} \oplus \overleftarrow{\textbf{h}^{j}_{t}}
\end{equation}
where $\textbf{w}^{j}_{t}$ is the word vector inputed into model at the $t$-th time step. $\overrightarrow{\textbf{h}^{j}_{t}}$ and $\overleftarrow{\textbf{h}^{j}_{t}}$ are hidden states generated by $\overrightarrow{GRU}$ and $\overleftarrow{GRU}$ respectively. $\oplus$ is the concatenation operation. We obtain the hidden states at each time step $[\textbf{h}^{j}_{1};\textbf{h}^{j}_{2};\ldots;\textbf{h}^{j}_{T_j}]$ and represent ${EDU}_j^i$ as $\textbf{x}_j^{\left(0\right)}$ through a max-pooling layer. In the following, we stack all the raw EDU vectors as the feature matrix
$\mathbf{X}^{\left(0\right)} \in R^{|U| \times m}=[\textbf{x}_1^{\left(0\right)};\textbf{x}_2^{\left(0\right)};\ldots;\textbf{x}_{|U|}^{\left(0\right)}]$\,, where $|U|$ represents the number of EDUs, $m$ denotes the dimension of EDU feature representation. We use this feature matrix $\mathbf{X}^{\left(0\right)}$ as the input for both the Sequence-based Representation Learning module and the Graph-based Representation Learning module.

\subsection{Sequence-based Representation Learning}  \label{Section4.3}
Different contexts can lead to diverse comprehensions of the text. In order to capture the rich contextual features in the EDU sequential structure, we use the TextCNN \citep{kim-2014-convolutional} to capture the important contextual relationship between locally co-occurring text units.

Given the raw feature matrix $\mathbf{X}^{\left(0\right)}\in R^{|U| \times m}$, 1D TextCNN convolutes adjacent EDUs through the fixed-size sliding window. Specifically, we define $m^\prime$ filters $w \in R^{k\times m}$, set each filter's window size as 3, and set the padding size as 1. These settings allow the model to take the context of an EDU into account when enhancing its representation. After sliding $m^\prime$ filters from the first EDU to the last EDU, the sequence-based representations for EDUs are obtained, denoted as the feature matrix $\textbf{X}_C\in R^{|U| \times m^\prime}$\,.

\subsection{Graph-based Representation Learning}  \label{Section4.4}

In order to deal with the EDU discourse dependency graph with multiple relations, we employ a Relation Graph Attention Network (RGAT) \citep{schlichtkrull2018modeling} to get the graph-based EDU representation. It can aggregate neighboring information according to the type of relation, and highlight key neighbor information with attention mechanism to fully grasp the internal relationship between nodes. 
 
We represent the discourse dependency graph generated in Section \ref{Section41} as $\mathcal{G}_i = \left( \mathcal{V}, \mathcal{E} \right)$\,, and take the matrix $\mathbf{X}^{\left(0\right)}\in R^{|U| \times m}$ as the original node feature matrix at the first layer. The representation of node $u$ could be $\textbf{x}_u^{\left(0\right)}\in R^m$. If node $u$ has $|R|$ kinds of edges connected to it, $\mathcal{N}_r^u$ denotes the set of the neighboring nodes of $u$ under the relation $r$, where $r \in \mathcal{R}$. As an example, given a specific node $u$, RGAT updates its representation with the following three steps: Firstly, different attention weights are learned and assigned to nodes connected with $u$ in neighboring set $\mathcal{N}_r^u$. Then, RGAT aggregates the neighbors' information according to their weights and gets the representation of node $u$ under relation $r$. Finally, all the representations of node $u$ under different types of relations are incorporated as the graph-based EDU representation contains rich multi-relation structural information.

Specifically, the feature representation of node $u$ in $0$ layer can be updated to $1$ layer as $\textbf{x}_u^{\left(1\right) }$:
\begin{equation} \label{eq1}
\textbf{x}_u^{\left( 1\right)}={ReLU}\left(\sum_{{r} \in \mathcal{R}} \sum_{{v} \in \mathcal{N}_{r}^{u}} \alpha_{{uv}}^{{r}} \textbf{W}^{{r}} \textbf{x}_{v}^{\left(0\right)}\right)	
\end{equation}
were, $\textbf{x}_{v}^{\left(0\right)}$ is the vector of neighboring node $v$ that connect to node $u$ under relation $r$ in the $0$ layer. $\textbf{W}^{{r}}$ is the parameter matrix for the particular relation type $r$. Here, ${ReLU}$ is the activate function, we use LeakyReLU. $\alpha_{{uv}}^{{r}}$ is used to measure the importance of neighbor node $v$ relative to node $u$ based on relation $r$.
\begin{equation}\label{eq2}
\alpha_{{uv}}^{{r}}=\frac{\exp \left(\textbf{W}^{{r}}\left(\textbf{x}_{{u}}^{(0)} \| \textbf{x}_{{v}}^{(0)}\right)\right)}{\sum_{{k} \in \mathcal{N}_{{r}}^{{u}}} \exp \left(\textbf{W}^{{r}}\left(\textbf{x}_{{u}}^{(0)} \| \textbf{x}_{{k}}^{(0)}\right)\right)}
\end{equation}

To alleviate over-parameterize problem, we use Basis Decomposition \citep{schlichtkrull2018modeling}. After this operation, we yield the enhanced vector for each EDU node at the 1 layer $[\textbf{x}_1^{\left(1\right)};\textbf{x}_2^{\left(1\right)};\ldots;\textbf{x}_{|U|}^{\left(1\right)}]$. Feature matrix $\textbf{X}_G\in R^{|U| \times m^\prime}$ denotes the graph-based EDU representations by stacking these EDU vectors, where $m^\prime$ is the dimensionality of output node embeddings.

Algorithm 1 shows the pseudocode of the Graph-based Representation Learning module.
\begin{algorithm*}
	\label{alg1}
	\caption{The algorithm for Graph-based Representation Learning} 
	\LinesNumbered  
	\KwIn{The Dependency discourse graph $\mathcal{G}_i = \left( \mathcal{V}, \mathcal{E} \right)$ \newline 
		The primary EDU nodes representations $\mathbf{X}^{\left(0\right)} \in R^{|U| \times m}=[\textbf{x}_1^{\left(0\right)};\textbf{x}_2^{\left(0\right)};\ldots;\textbf{x}_{|U|}^{\left(0\right)}]$ 
	}
	\KwOut{The graph-based EDU representations $\textbf{X}_G\in R^{|U| \times m^\prime}$
}
	Get the primary feature representation of EDU node $u$ at layer $0$ as $\textbf{x}_{u}^{\left(0\right)}$\,.\newline
	\ForEach{type of relation $r \in \mathcal{R}$ connected to EDU node $u$}{
		Get the neighboring nodes set $\mathcal{N}_r^u$ of node $u$ under the relation $r$\,.\newline
		\ForEach {neighboring node $v \in \mathcal{N}_r^u$} {Calculate the weight value $\alpha_{{uv}}^{{r}}$ of neighboring node $v$ relative to node $u$ by Equation \ref{eq2}.
		}
    Aggregating all the neighboring nodes according to their weight values, and obtain the representation of node $u$ under relation $r$.
	}
	Sum the representations of node $u$ under all types of relations, and get the updated node representation $\textbf{x}_{u}^{\left(1\right)}$ in layer 1 through a ${ReLU}$ activate function.\newline
	\textbf{repeat} above calculation steps until all EDU nodes in set $\mathcal{V}$ are updated.\newline
	\Return the graph-based EDU representations $\textbf{X}_G\in R^{|U| \times m^\prime}=[\textbf{x}_1^{\left(1\right)};\textbf{x}_2^{\left(1\right)};\ldots;\textbf{x}_{|U|}^{\left(1\right)}] $
	
\end{algorithm*}  
 
 Finally, the representation $\textbf{X}_{GC} \in R^{|U| \times 2m^\prime}$ of EDU is the concatenation of the sequence-based representation $\textbf{X}_C$\,(Section \ref{Section4.3}) and the graph-based representation $\textbf{X}_G$\,(Section \ref{Section4.4}), where $2m^\prime$ is the dimension.  

\subsection{GRU-GA-based Text Representation}  \label{Section4.5}
To fuse all the text unit representations and form the final text representation for prediction, we design a fusion network named Gated Recursive Unit combined with Global Attention mechanism (GRU-GA), which highlights the important EDU while integrating the entire text information. The GRU network \citep{cho-etal-2014-learning} sequentially re-learns the enhanced representation of the EDUs in the top-down global reading order. Assuming the network has $T$ time steps, it inputs an enhanced EDU representation of the feature matrix $\textbf{X}_{GC}\in R^{|U| \times 2m^\prime}$ at each time step, and obtains the hidden states $[\textbf{h}_{1};\textbf{h}_{2};\ldots;\textbf{h}_{T}]$.

Consider that not all hidden states have the same contribution for the resulting text representation, we use the global attention \citep{luong-etal-2015-effective} to compute the weight of each hidden state. For the hidden state $\textbf{h}_{t}$ at time step $t$, its weight $\alpha_{t}$ is calculated from the current state $\textbf{h}_{t}$ and the state $\textbf{h}_{T}$ output at the last time step. Formally:
\begin{equation}
\alpha_{t}=\frac{\exp \left(\textbf{h}_{T}^{T} \textbf{h}_{t}\right)}{\sum_{t^{\prime}=1}^T \exp \left(\textbf{h}_{T}^{T} \textbf{h}_{t^{\prime}}\right)}
\end{equation}

The text representation $\textbf{z}$ is calculated as the weighted average of all the hidden states.
\begin{equation}
\textbf{z}=\sum_{t=1}^{T}\alpha_{t}\textbf{h}_t
\end{equation}
\subsection{Inductive Classification}  \label{Section4.6}
The inductive classification method learns inductive patterns from existing data and applies them to new data. Same as \cite{zhang-etal-2020-every}, this paper uses inductive classification method and each text is an individual graph for whole graph classification. We use a fully connected layer with Softmax activation function to map the text representation $\textbf{z}$ to the probability values. 

\begin{equation}
\hat{y} = Softmax \left( \mathbf{W}_y \mathbf{z} + \mathbf{b}_y \right)
\end{equation}
here, $\hat{y}$ is the predicted probability. $\mathbf{W}_y$ is a weight metric and $\mathbf{b}_y$ is a bias vector. The cross-entropy loss function is defined as:
\begin{equation}
\mathcal{L} = - y \log \hat{y}-\left( 1 - y \right) \log \left( 1 -\hat{y} \right)
\end{equation}   
$y \in \left\{ 0, 1 \right\}$ is the ground-truth label of the input text. 

\section{Experiments}
We mainly answer the following evaluation questions to evaluate the effectiveness of EDU4FD:
\begin{itemize}
	\item[\textbf{EQ1}] Does EDU4FD perform better than the state-of-the-art comparative models on cross-source datasets?
	\item[\textbf{EQ2}] How effective are the Sequence-based Representation Learning module, the Graph-based Representation Learning module and the GRU-GA-based Text Representation module in improving the fake news detection ability of EDU4FD?
	\item[\textbf{EQ3}] Can EDU4FD provide reasonable explanation about the fake news detection results?
	\item[\textbf{EQ4}] Does EDU4FD show the high-quality representation over other methods in visualization study?	
\end{itemize}
\subsection{Dataset Description}
Due to over-fitting, previous algorithms usually cannot generalize to new texts from new source that are not seen in the training set. However, fake news published from different sources varies greatly in style. Models must classify news from different sources to reduce the over-reliance on corpus. Therefore, we conducted experiments on two cross-source datasets groups: (1) LUN and SLN. (2) Kaggle, BuzzFeed, and PolitiFact. 

(1) LUN and SLN

\textbf{LUN}\footnote{The LUN dataset could be obtained from https://homes.cs.washington.edu/~hrashkin/fact$\_$checking$\_$files/.}\textbf{:} LUN is a well-known fake news dataset obtained from \cite{rashkin-etal-2017-truth}. News articles in LUN are divided into two sub-datasets, LUN-train and LUN-test, depending on the source of publication. The LUN-train dataset contains news from the Onion and the Gigaword news excluding 'APW'\footnote{'APW' is the abbreviation of 'Associated Press Worldstream'} and 'WPB'\footnote{'WPB' is the abbreviation of 'Washington Post/Bloomberg Newswire service'}, while the LUN-test dataset covers the rest of the Gigaword news resources (only 'APW' and 'WPB' sources). 

\textbf{SLN}\footnote{The SLN dataset could be obtained from http://victoriarubin.fims.uwo.ca/news-verification/data-to-go/.}\textbf{:}
The SLN dataset is a widely used dataset for fake news detection \citep{rubin2016fake}. It contains news sources from the Toronto Star, the NY Times, the Onion and the Beaverton sources. 

For the sake of consistency, we followed \cite{WANG2021114090} and took LUN and SLN datasets as the cross-sources datasets. Specifically, we used LUN-train as the training dataset and set LUN-test and SLN dataset as its two test datasets. The two test datasets mentioned above can be called cross-sources test sets relative to LUN-train because the style of news articles contained in the test sets are completely different from the training set. This setting can better detect the generalization ability of the model. We summarize the statistics of three datasets in Table \ref{tab:dataset1}.

(2) Kaggle, BuzzFeed, and PolitiFact

\textbf{Kaggle}\footnote{The Kaggle dataset could be obtained from https://www.kaggle.com/jruvika/fake-news-detection.}\textbf{:} Kaggle is a publicly available fake news dataset consists of 4009 news, with 1872 labeled real and 2137 labeled fake. We obtained this dataset from kaggle.com.

\textbf{BuzzFeed\footnote{The BuzzFeed and PolitiFact datasets could be obtained from https://github.com/KaiDMML/FakeNewsNet.}:} This dataset is compiled by \cite{shu2017exploiting}. It contains news headlines and news bodies on Facebook. In this paper, we only utilized the news body content.

\textbf{PolitiFact}\footnote{http://www.politifact.com/}\textbf{:} This dataset is also obtained from \cite{shu2017exploiting}. It is collected from well-recognized fact-checking website politifact.com.

Similar to the previous setting, we treated Kaggle datasets as the training set and used BuzzFeed and PolitiFact datasets as its two cross-sources test datasets. The two test sets and the training set were collected from different sources.
Tabel \ref{tab:dataset2} shows the statistics of these three datasets.
\begin{table*}
	\centering
	\caption{Descriptive statistics of the LUN-train, LUN-test and SLN datasets}
	\label{tab:dataset1}
	\begin{tabular}{llll}
		\hline
		Statistic  & LUN-train & LUN-test& SLN \\
		\hline
		$\#$ Real news      & 9,995  & 750& 180     \\
		$\#$ Fake news       & 14,047  & 750&180     \\
		$\#$ Total news  &24,042& 1500  &360\\
		avg.$\#$ EDUs per news &42.50&45.16&62.50\\
		\hline
	\end{tabular}
	
\end{table*}

\begin{table*}
	\centering
	\caption{Descriptive statistics of the Kaggle, BuzzFeed and PolitiFact datasets}
	\label{tab:dataset2}
	\begin{tabular}{llll}
		\hline
		Statistic  &Kaggle &BuzzFeed& PolitiFact \\
		\hline
		$\#$ Real news    & 1872 & 90&111      \\
		$\#$ Fake news    & 2137  & 80&92     \\
		$\#$ Total news   & 4009 &170&203    \\
		avg.$\#$ EDUs per news &53.37&57.61&56.79\\
		\hline
	\end{tabular}
	
\end{table*}
\subsection{Comparison Methods}
We compared our EDU4FD model against several strong baselines, which can be divided into 3 categories: 

(1) Traditional machine learning methods
\begin{itemize}
	\item \textbf{SVM} \citep{10.5555/559923}: A support vector machine classifier with the linear kernel is utilized to detect fake news. Here, we employed Term Frequency-Inverse Document Frequency (TF-IDF) and got term frequency values of n-grams vocabulary features as the input features.
	\item  \textbf{Logistic Regression} \citep{kleinbaum2002logistic}: The logistic regression classifier uses text characteristics vectorized by the TF-IDF method to detect fake news. 
\end{itemize}

(2) Non-graph deep learning network methods
\begin{itemize}
	\item \textbf{CNN} \citep{kim-2014-convolutional}: The Convolutional Neural Network utilizes a 1-d convolution layer with filters of size 3, followed by a max-pooling layer and a fully connected layer to detect fake news.
	\item \textbf{BiGRU} \citep{cho2014learning}: The bidirectional GRU network is based on a pair of bidirectional GRU layers, which could capture context information. First, the low-dimensional word vectors are inputed into the BiGRU network. Then, the text representation is learned for fake news detection.
	\item \textbf{BERT} \citep{devlin-etal-2019-bert}: The Google-BERT is a state-of-the-art pre-trained model. We first vectorized each sentence with BERT and then fed them into the LSTM network to classify whether the news is fake or real. 
\end{itemize} 

(3) Graph-based deep learning network methods
\begin{itemize}
	\item \textbf{GCN} \citep{vaibhav-etal-2019-sentence}: This method applies graph convolutional network (GCN) \citep{DBLP:conf/iclr/KipfW17} on the complete graph in which sentences are fully connected, and the adjacency matrix take the form of all 1 with 0 on the diagonal. The method benefits from this structure as it could capture the long-distance dependencies between sentences in the text. Specifically, it first encodes sentences in the text by an LSTM network. And then a GCN is applied to enhance the sentence representation. Finally, the text representation is obtained by a max-pooling layer and a fully connected layer to detect fake news.
	\item \textbf{GAT} \citep{vaibhav-etal-2019-sentence}: A Graph Attention Network (GAT) \citep{velivckovic2018graph} is utilized to learn the representation of news text. This method also first fully connects sentences within the text as a graph and then applies a GAT network. When learning sentence representation, it aggregates neighboring features according to different weights.
	\item \textbf{GAT2H} \citep{vaibhav-etal-2019-sentence}: This model learns text representation using Graph Attention Network with two attention heads (GAT2H). GAT2H is applied on the same fully connection graph as above, and the output of each attention head are concatenated and then fed into the classification layer.
	\item \textbf{SemSeq4FD} \citep{WANG2021114090}:  SemSeq4FD is a novel graph-based neural network model for fake news detection. It considers the global semantic relations feature, local sequential order feature, and the global sequential order feature among sentences simultaneously. In SemSeq4FD, a complete graph is built by fully connecting sentences. It utilizes a graph convolutional network with self-attention mechanism and a TextCNN to learn the enhanced sentence representation. Finally, it uses an LSTM network to integrate text representation for fake news detection.
\end{itemize} 

\subsection{Experimental Setup} 
\textbf{Environment: }The experimental environment is Intel i7 2.20 GHz processor, 8.0 GB memory, GTX-1050 ti GPUs. All the deep learning network methods in baselines are implemented with Pytorch libraries (1.1.0). 

\textbf{Data Preprocessing: }Same as \cite{vaibhav-etal-2019-sentence}, we first randomly took out 10\% of the entire dataset for test. We then randomly divided the rest of the dataset into 80\% training and 20\% validation subsets. We preprocessed the articles with the following
rules: First, we segmented the news text as EDUs with rhetoric relations (Section \ref{Section41}). Distributions of 19 kinds of relations in all datasets are shown in Appendix \ref{appendix}. Then we removed the news that has less than $2$ EDUs.

\textbf{Hyperparameters: }For the sake of consistency, we followed \cite{WANG2021114090} to set the same hyperparameters. We set the optimized learning rate as $10^{-3}$, the dropout rate as $0.2$, the size of each batch as $32$, and the number of epoch as $10$. The threshold to control the maximum length of each EDU is $200$.

\textbf{Evaluation Metrics: }We adopt the general evaluation criteria of text classification, including accuracy, precision, recall, and F1 Score. All standards use macro-average calculation. All results on all datasets have been averaged over 5 trials. 
\subsection{Performance Comparison (\textbf{EQ1})}

\begin{table*}
	\centering
	\caption{Experimental results on LUN-test and SLN. Results marked $*$ taken from \cite{WANG2021114090}}
	\resizebox{\textwidth}{24mm}{
		\renewcommand{\arraystretch}{1}
		\begin{tabular}{llllllllllll}
			\toprule
			Datasets  & Metric & SVM$^*$ & LR$^*$ & CNN$^*$ & BiGRU & BERT$^*$ & GCN$^*$ & GAT$^*$ & GAT2H$^*$ & SemSeq4FD$^*$ & \textbf{EDU4FD} \\ 
			\bottomrule
			\multirow{4}*{LUN-test} & Accuracy& 0.7886
			&0.7893
			&0.9094
			&0.8904
			&0.8346
			&0.9224
			&0.9255
			&0.9178
			&\underline{0.9378} & \textbf{0.9591}  \\
			& Precision & 0.8105
			&0.8083
			&0.9112
			&0.8949
			&0.8356
			&0.9248
			&0.9281
			&0.9212
			&\underline{0.9390}
			&\textbf{0.9597}
			\\
			& Recall &0.7886
			&0.7893
			&0.9088
			&0.8909
			&0.8346
			&0.9222
			&0.9251
			&0.9178
			&\underline{0.9378}
			&\textbf{0.9593}
			\\
			& F1 Score  &0.7848
			&0.7860
			&0.9086
			&0.8902
			&0.8345
			&0.9222
			&0.9251
			&0.9176
			&\underline{0.9378}
			&\textbf{0.9591}
			\\
			\bottomrule
			\multirow{4}{*}{SLN}& Accuracy& 0.8333
			&0.8388
			&0.6452
			&0.7950
			&0.7583
			&0.8640
			&0.8538
			&0.8584
			&\underline{0.8842}
			&\textbf{0.8939}
			\\
			& Precision & 0.8337
			&0.8390
			&0.6466
			&0.7961
			&0.7662
			&0.8670
			&0.8567
			&0.8600
			&\underline{0.8904}
			&\textbf{0.8953}
			\\
			& Recall &0.8333
			&0.8388
			&0.6452
			&0.7950
			&0.7583
			&0.8640
			&0.8538
			&0.8584
			&\underline{0.8842}
			&\textbf{0.8939}
			\\
			& F1 Score 
			&0.8332
			&0.8388
			&0.6440
			&0.7948
			&0.7565
			&0.8638
			&0.8535
			&0.8580
			&\underline{0.8838}
			&\textbf{0.8938}
			\\
			\bottomrule
	\end{tabular}}	
	\label{tab:dataset3}
\end{table*}
\begin{table*}
	\centering
	\caption{Experimental results on BuzzFeed and PolitiFact}
	\resizebox{\textwidth}{24mm}{
		\renewcommand{\arraystretch}{1}
		\begin{tabular}{llllllllllll}
			\toprule
			Datasets  & Metric & SVM & LR & CNN & BiGRU & BERT & GCN & GAT & GAT2H & SemSeq4FD & \textbf{EDU4FD} \\ 
			\hline
			\multirow{4}*{BuzzFeed} & Accuracy& 0.6547
			&0.6309
			&0.6559
			&0.6107
			&0.6428
			&0.6083
			&0.6000
			&0.6249
			&\underline{0.7024}
			&\textbf{0.7488}
			\\
			& Precision & 0.6568
			&0.6299
			&0.6940
			&0.6323
			&0.6540
			&0.6458
			&0.6357
			&0.6625
			&\underline{0.7113}
			&\textbf{0.7519}
			\\
			& Recall & 0.6500
			&0.6295
			&0.6443
			&0.5990
			&0.6346
			&0.5935
			&0.5850
			&0.6120
			&\underline{0.6969}
			&\textbf{0.7486}
			\\
			& F1 Score  & 0.6487
			&0.6296
			&0.6275
			&0.5784
			&0.6276
			&0.5565
			&0.5443
			&0.5843
			&\underline{0.6942}
			&\textbf{0.7475}
			\\
			\hline
			\multirow{4}{*}{PolitiFact}& Accuracy& 0.6464
			&0.6262
			&0.6010
			&0.5761
			&0.5858
			&0.6293
			&0.6343
			&0.6384
			&\underline{0.6614}
			&\textbf{0.7162}
			\\
			& Precision
			&0.6524
			&0.6310
			&0.5977
			&0.5670
			&0.5819
			&0.6759
			&\underline{0.6774}
			&0.6723
			&0.6527
			&\textbf{0.7155}
			\\
			& Recall & 0.6263
			&0.6038
			&0.5786
			&0.5499
			&0.5599
			&0.5960
			&0.6029
			&0.6084
			&\underline{0.6384}
			&\textbf{0.7111}
			\\
			& F1 Score & 0.6202
			&0.5937
			&0.5652
			&0.5297
			&0.5408
			&0.5630
			&0.5722
			&0.5819
			&\underline{0.6255}
			&\textbf{0.7110}
			\\
			\bottomrule
	\end{tabular}}
	\label{tab:dataset4}
\end{table*}
We compared EDU4FD with 9 different baselines on four cross-source datasets, as shown in Table \ref{tab:dataset3} and Table \ref{tab:dataset4}. We \underline{underlined} the best baseline results, and \textbf{bold} the best experimental results. The results marked with $*$ in Table \ref{tab:dataset3} are taken from \cite{WANG2021114090}. The following conclusions can be drawn from the observation of Table \ref{tab:dataset3} and Table \ref{tab:dataset4}.

(1) Compared with all baselines, EDU4FD achieves the most advanced results on four cross-source test sets, and improves F1 values by 2.13\%, 1\%, 5.33\%, and 8.55\% respectively compared with the best baselines, suggesting that our model is more generalized and robust than others. This also shows the effectiveness of modeling structure information from the perspective of EDU. The discourse structure we utilized could benefits the classification model. 

(2) It is worthwhile to point out that EDU4FD is better than all graph-based deep learning methods. EDU4FD applies the dependency graph structure, which abandons the defect that graph-based baselines only focus on sentences and cannot clearly understand the meaningful relationship in the text. 
\subsection{Ablation Analysis (\textbf{EQ2})}
In order to answer EQ2, five variants of EDU4FD models are designed, which remove part of the whole model to explore the validity of these parts.
\begin{itemize}
	\item \textbf{EDU4FD$\backslash$EDU}: This model is a variant of EDU4FD that eliminates EDU. The validity of EDU is verified by replacing the input of EDU4FD with sentences. The RGAT network is replaced by GAT, and the EDU dependency graph is replaced by the fully connected graph. 
	\item \textbf{EDU4FD$\backslash$RGAT}: This model is a variant of EDU4FD, which does not consider the dependency graph. It removes the Graph-based Representation Learning module and only sequence information affects the model.
	\item \textbf{EDU4FD$\backslash$C}: This model is a variant of EDU4FD, which excludes the Sequence-based EDU Representation Learning module and does not learn the coherence and consistency between adjacent EDUs in local order. 
	\item \textbf{EDU4FD$\backslash$G}: This model is a variant of EDU4FD, which does not use a fusion network (GRU-GA) to integrate the text representation in global order. The GRU-GA-based Text Representation module is replaced by a max-pooling layer.
	\item \textbf{EDU4FD$\backslash$C$\backslash$G}: This model is a variant of EDU4FD, which excludes both the Sequence-based EDU Representation Learning module and the GRU-GA-based Text Representation module. We used this variant to validate the impact of modeled sequence information. The two modules we eliminated could model local order and global order information respectively. After encoding EDUs, we only inputed EDUs representations to the RGAT network and learn the graph-based EDU representation. The outputs are fed into a max-pooling layer for classification. 
\end{itemize}
\begin{figure*}[htbp]
	\centering
	\subfloat[LUN-test]{
		\includegraphics[width=0.46\linewidth,height=18em]{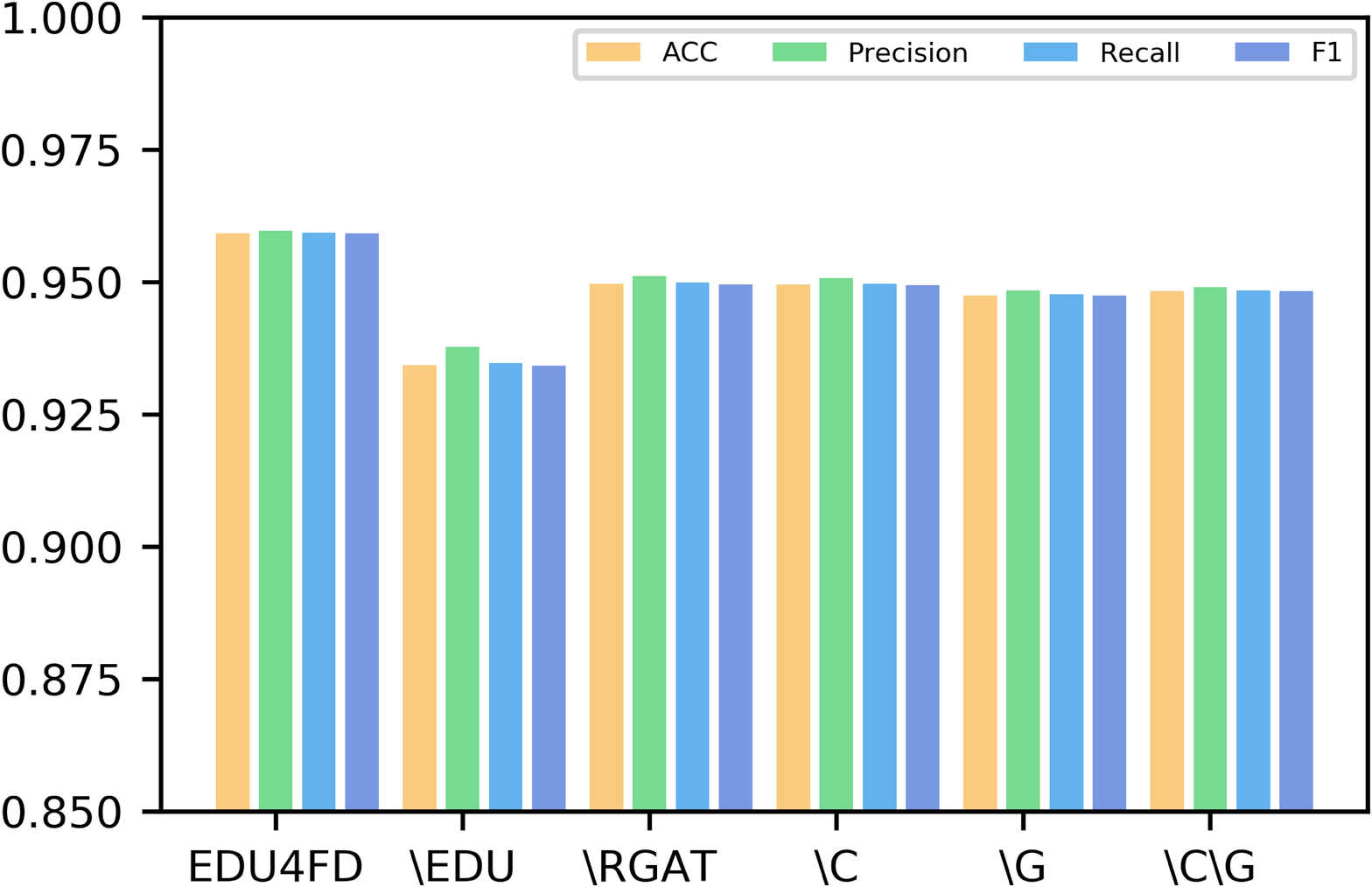}
		\label{fig:ablation1}
	}
	\quad
	\subfloat[SLN]{
		\includegraphics[width=0.46\linewidth,height=18em]{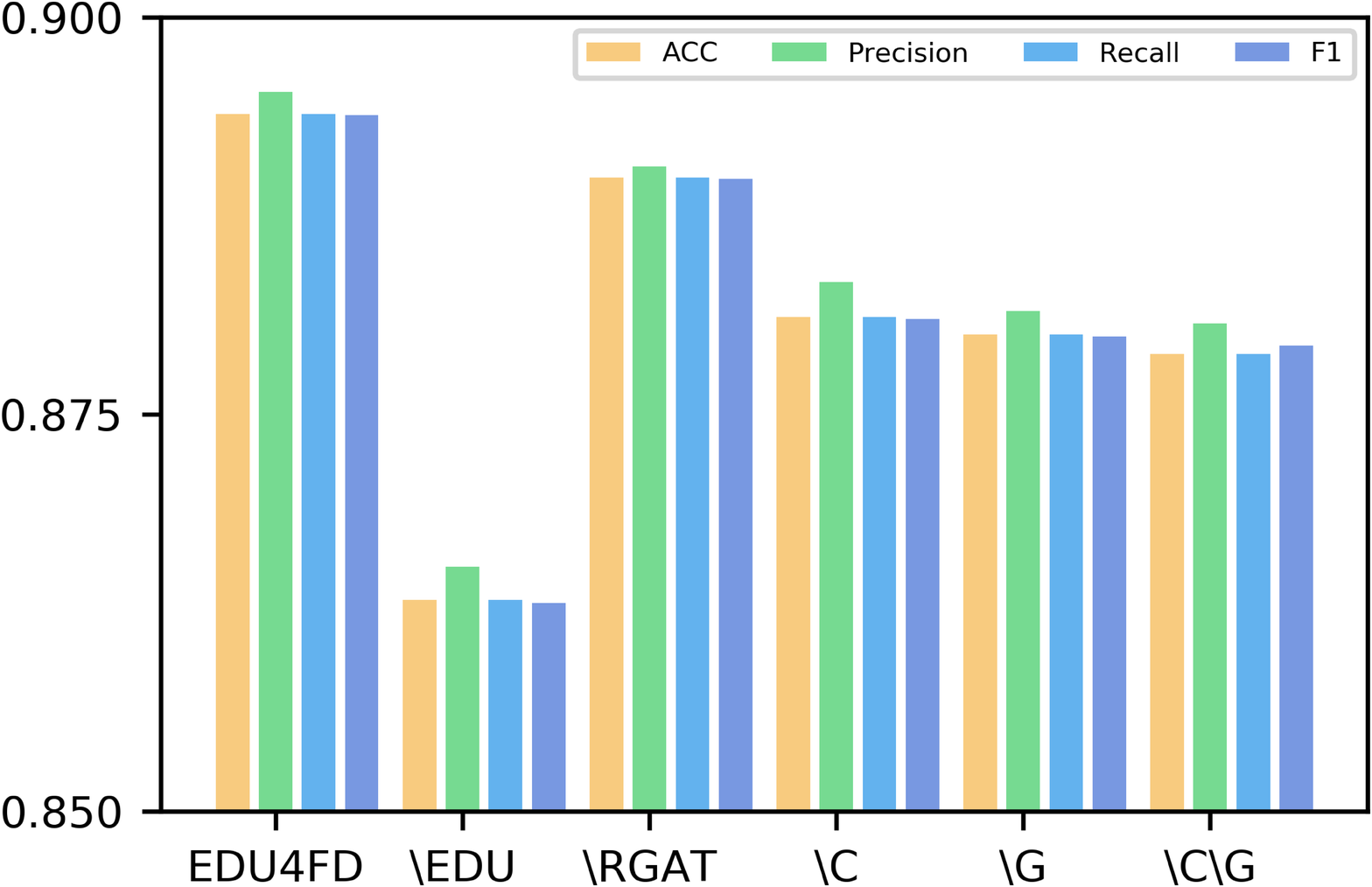}
		\label{fig:ablation2}
	}
	
	\subfloat[BuzzFeed]{
		\includegraphics[width=0.46\linewidth,height=18em]{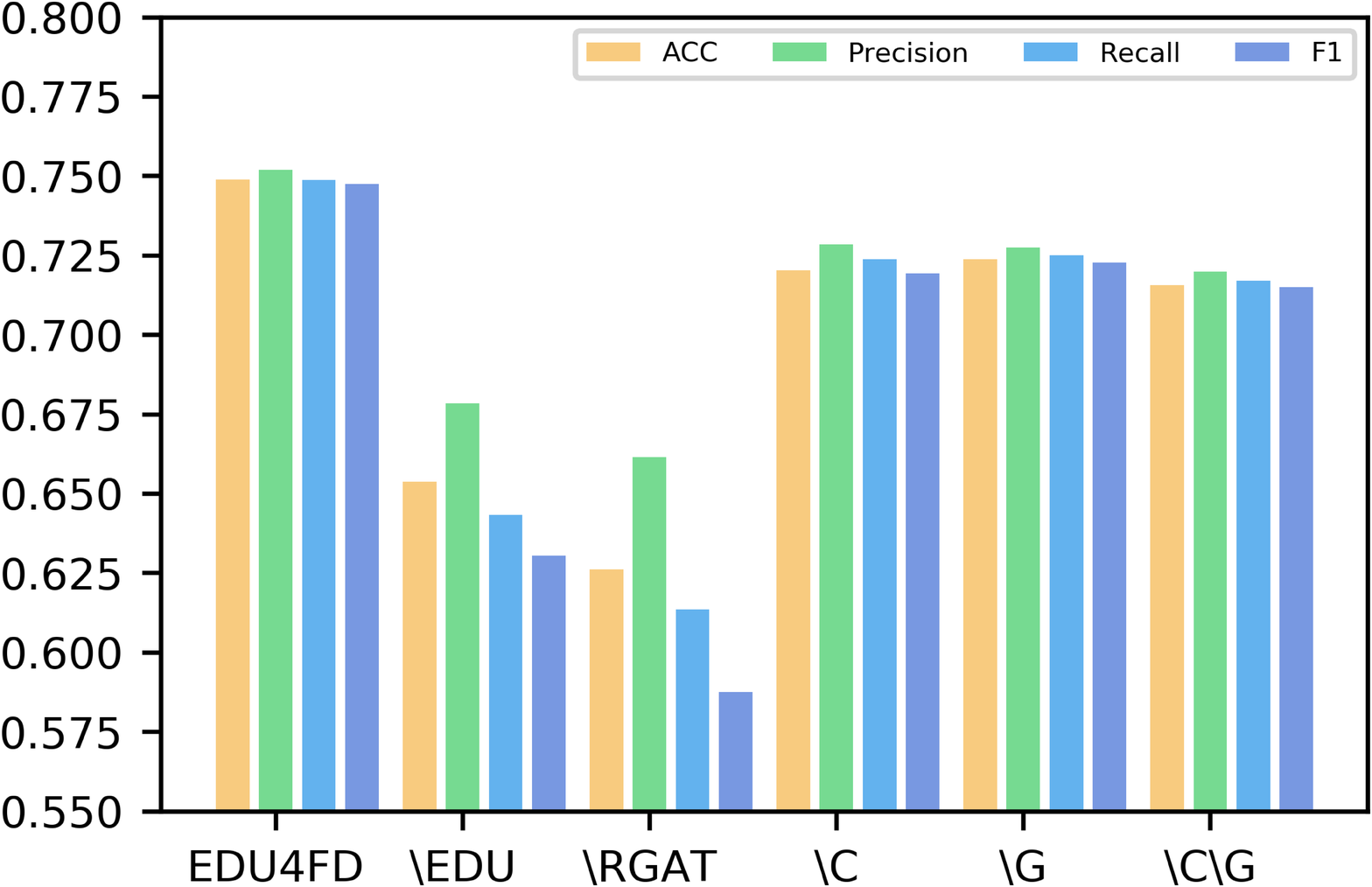}
		\label{fig:ablation3}
	}
	\quad
	\subfloat[PolitiFact]{
		\includegraphics[width=0.46\linewidth,height=18em]{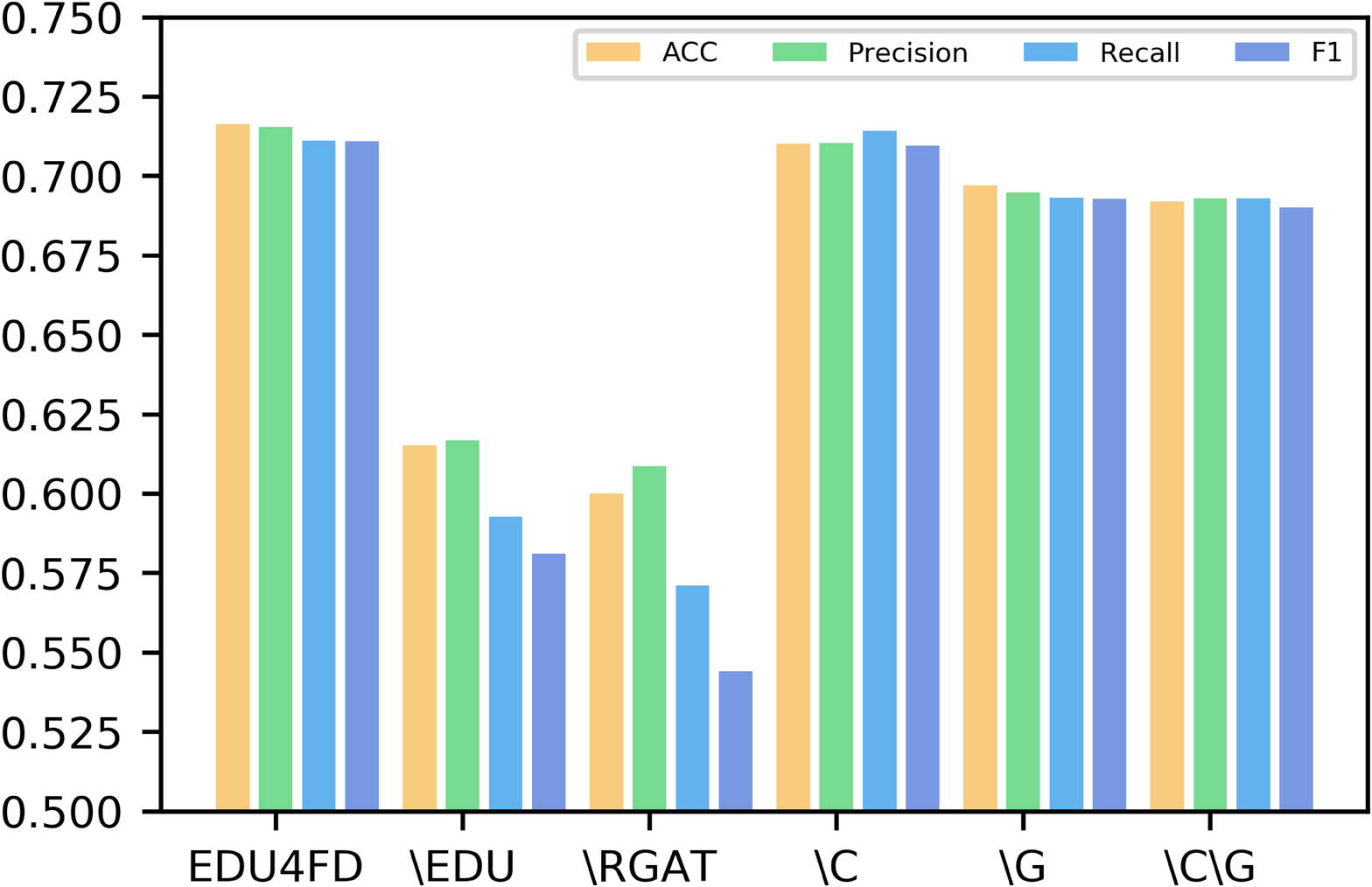}
		\label{fig:ablation4}
	}
	\caption{Ablation results of EDU4FD on four test sets}
\end{figure*}

The performances of variants on four data sets are shown in Figure \ref{fig:ablation1}, Figure \ref{fig:ablation2}, Figure \ref{fig:ablation3}, and Figure \ref{fig:ablation4}, from which we could find that:

\begin{itemize}
	\item The effectiveness of EDU4FD has reduced after removing any modules, which demonstrates the rationality and validity of the design of the EDU4FD model.
	\item EDUs could express more coherent information than words and more specific information than sentences. Hence, using EDUs instead of words or sentences plays an important role in effective fake news detection.
	\item When we remove the Graph-based Representation Learning module, the performance of EDU4FD$\backslash$RGAT degrades sharply in comparison to EDU4FD in most of datasets. It suggests that the usage of rhetorical relations is necessary. In particular, we can capture more structural relationships and assign attention weights to important EDU nodes by the relation graph attention network.
	\item When we disregard the sequence between consecutive EDUs, in contrast to EDU4FD in terms of Accuracy and F1 Score, EDU4FD$\backslash$C's performance reduced by 0.96\% and 0.97\% on LUN-test, 1.28\% and 1.28\% on SLN, 2.86\% and 2.83\% on BuzzFeed, and 0.61\% and 0.16\% on PolitiFact. It verifies the substantial influence of modeling coherence relationship between consecutive EDUs with TextCNN.
	\item Compared to EDU4FD, the performance of EDU4FD$\backslash$G reduced by 1.17\% and 1.17\% on LUN-test, 1.39\% and 1.39\% on SLN, 2.50\% and 2.47\% on BuzzFeed, and 1.92\% and 1.83\% on PolitiFact, comparing against the best results in terms of Accuracy and F1 Score. The results suggest that the global attention mechanism network could enable the model to focus on key EDUs, and it is helpful for understanding the text in top-down global reading order.
	\item When we eliminate both the Sequence-based Representation module and the GRU-GA-based Text Representation module, compared with EDU4FD, the performance of EDU4FD$\backslash$C$\backslash$G degrades on four data sets. Therefore, both local order and global order contribute to the performance. The combination of the two modules guides the superiority of the model.
\end{itemize}
\subsection{Case Study (\textbf{EQ3})}
\begin{figure*}[h]
	\centering
	\includegraphics[width=0.9\linewidth, height=20em]{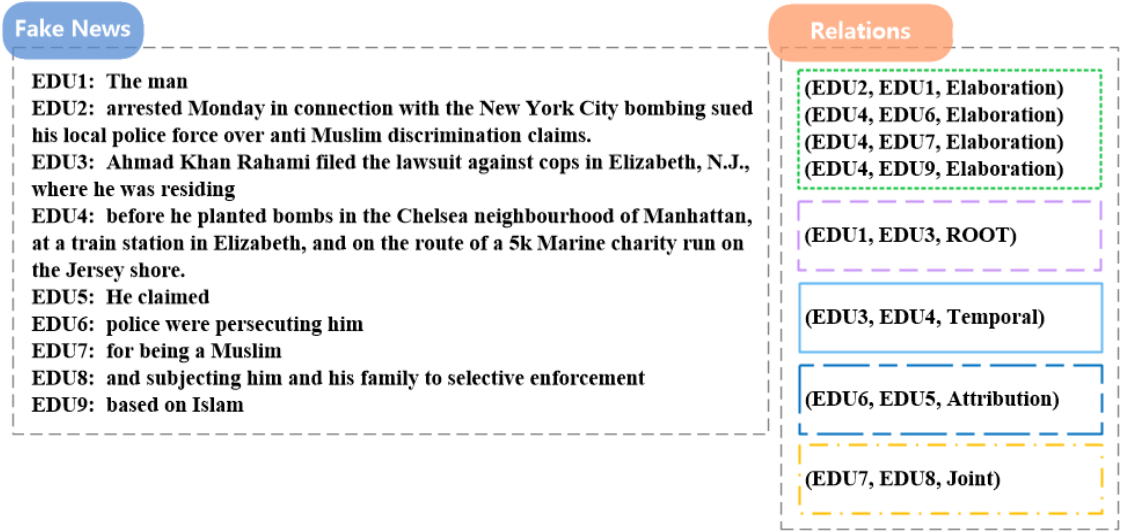}
	\caption{The explainable relations of a text captured by EDU4FD}
	\label{fig:case0}
\end{figure*}

\begin{figure*}[h]
	\centering
	\includegraphics[width=0.8\linewidth,height=26em]{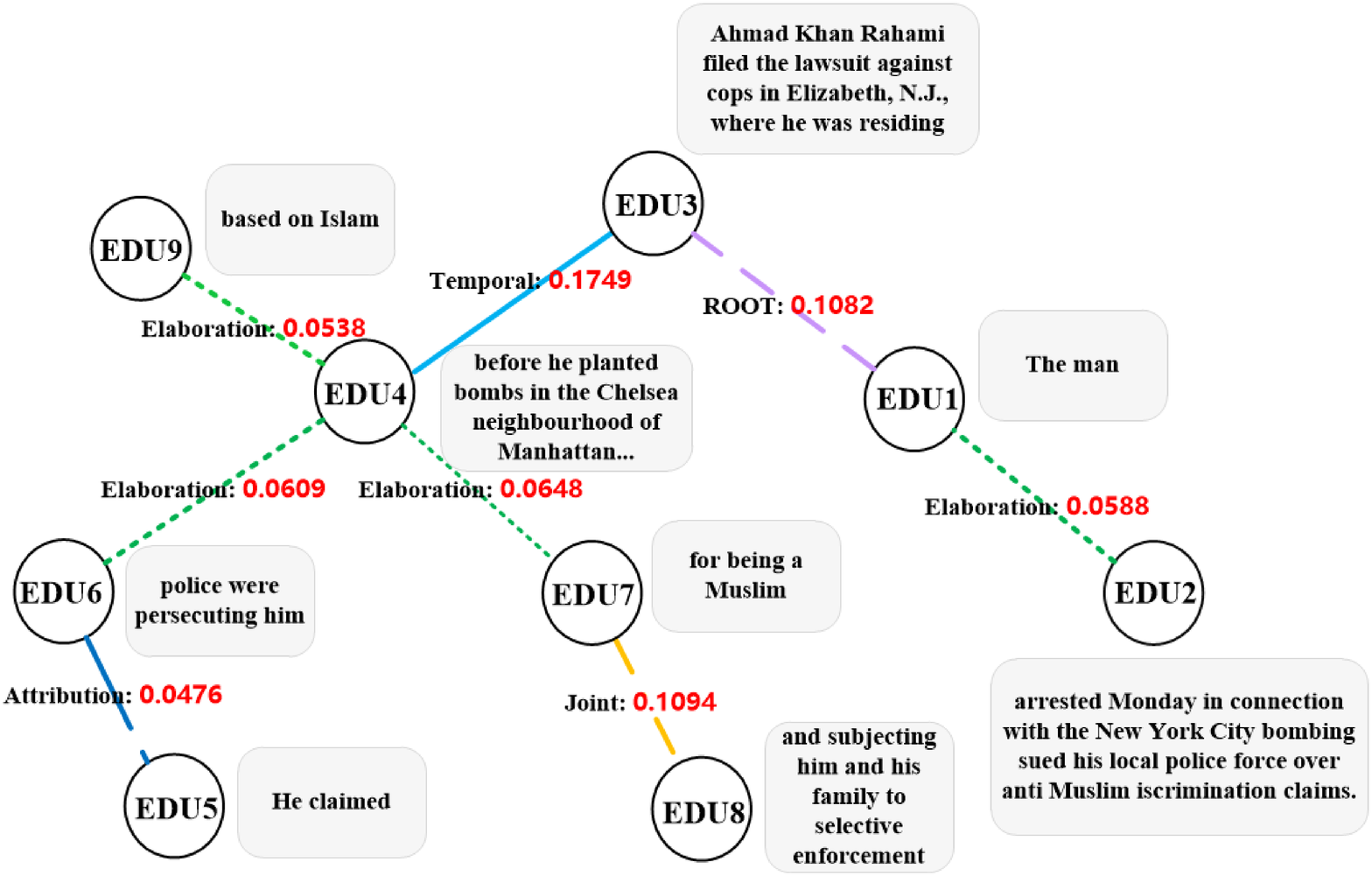}
	\caption{The visualization with attention weights of a text captured by EDU4FD}
	\label{fig:case1}
\end{figure*}

To answer EQ3, we used an example to illustrate the important role of functional rhetorical relations and the relation graph attention network in improving the explanatory ability of our model. We randomly selected an example of fake news from the BuzzFeed dataset. Figure \ref{fig:case0} shows the EDUs of the news text (the original text has been segmented to EDUs) and the corresponding rhetorical relation between each pair of EDUs. Figure \ref{fig:case1} shows the attention weights between EDUs captured by EDU4FD.

As we can see, the connection between $EDU3$ and $EDU4$ has the highest attention weight (use bold solid line in Figure \ref{fig:case1}). In contrast, other neighbors of $EDU4$ with the Elaboration relationship have lower attention scores than $EDU3$. It can be assumed that when enhancing $EDU4$'s node representation, the model relies more on the neighboring node under Temporal relationship. Hence, our EDU4FD can express structural information and provide useful visual hints for fake news classification.
\subsection{Text Representation Visualization (\textbf{EQ4})}
\begin{figure*}[h]
	\centering
	\subfloat[BERT]{
		\includegraphics[width=0.33\linewidth,height=15em]{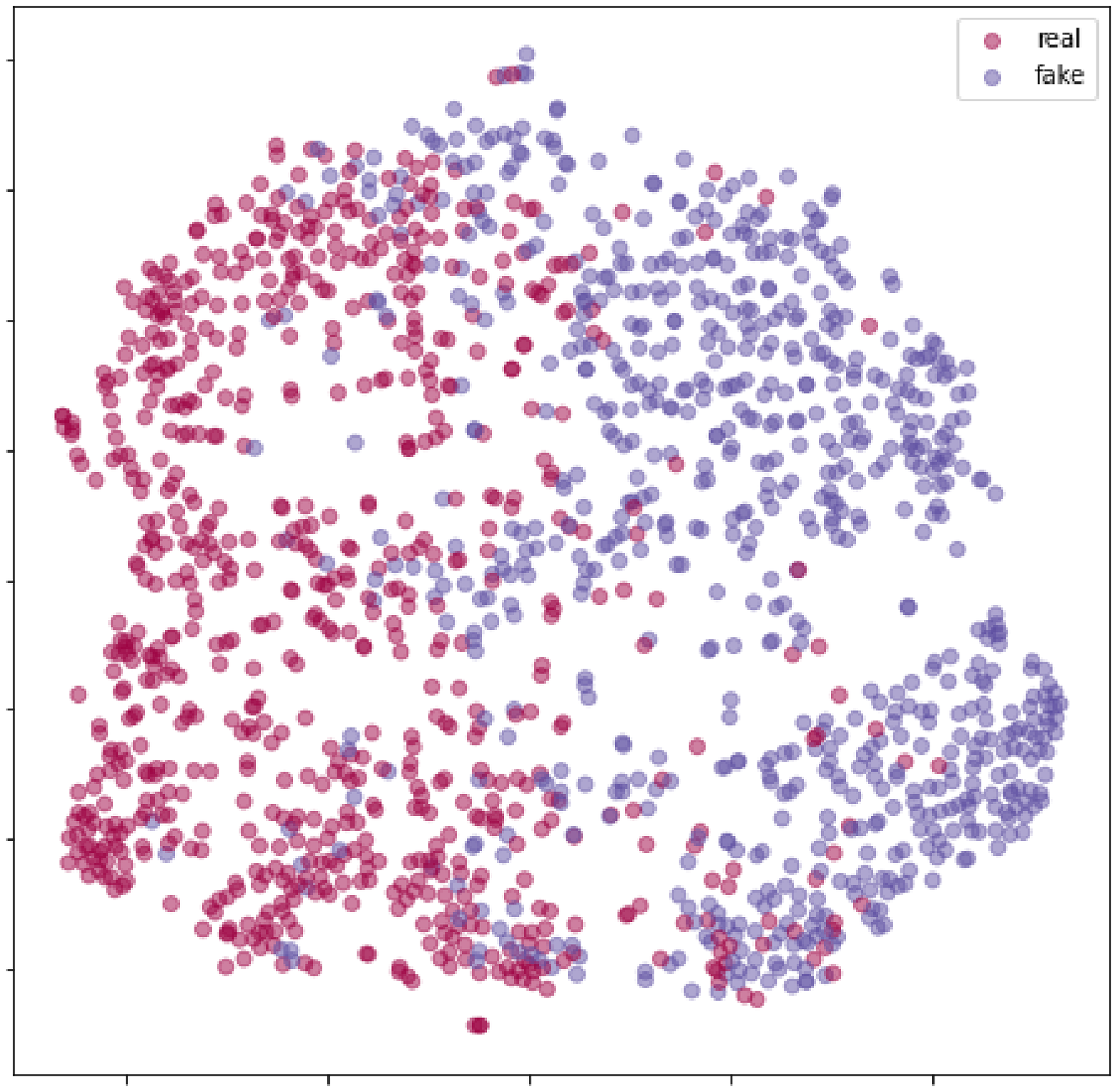}
		\label{fig:bert}
	}
	\subfloat[GCN]{
		\includegraphics[width=0.33\linewidth,height=15em]{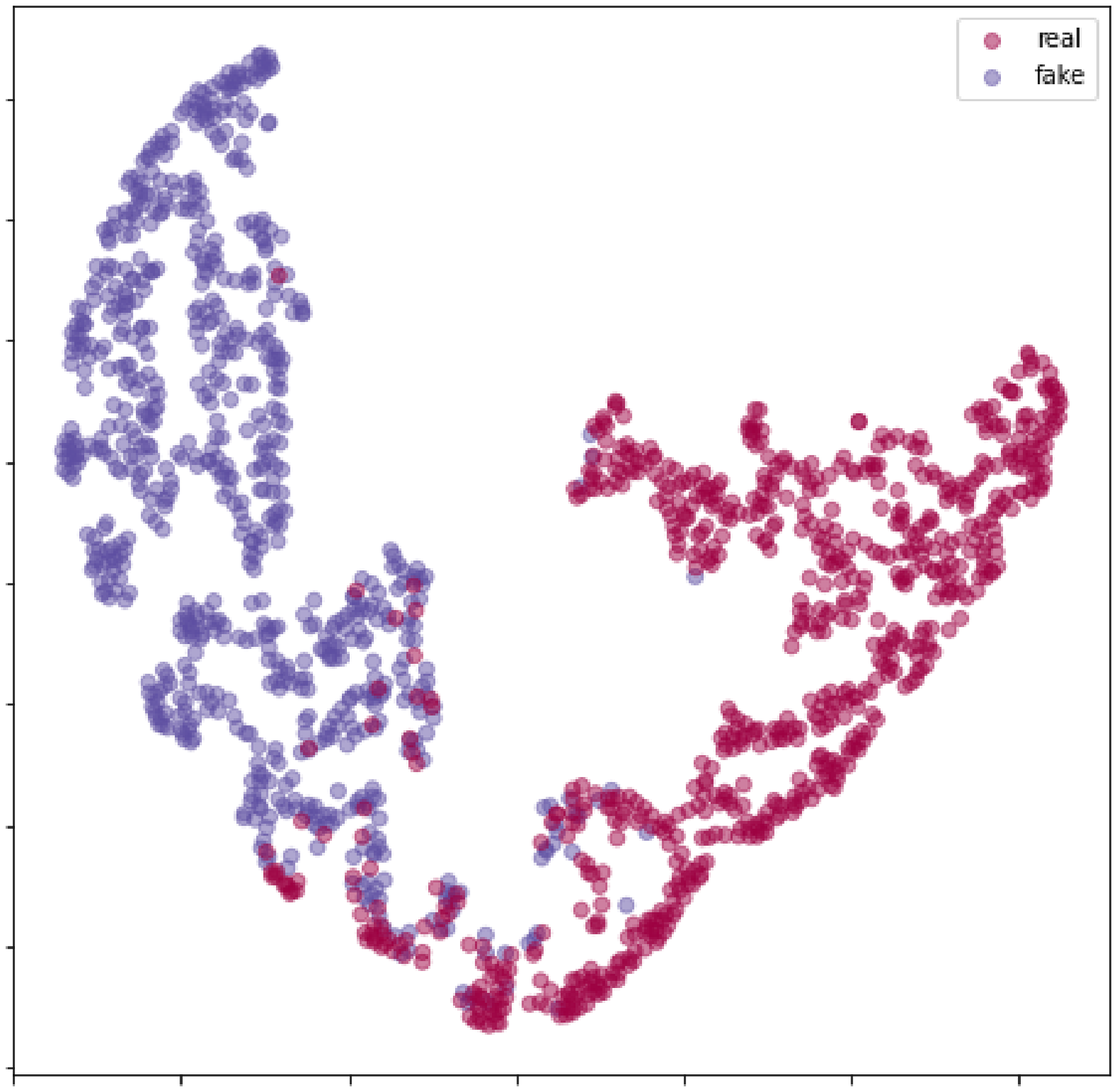}
		\label{fig:gcn}
	}
	\subfloat[GAT]{
		\includegraphics[width=0.33\linewidth,height=15em]{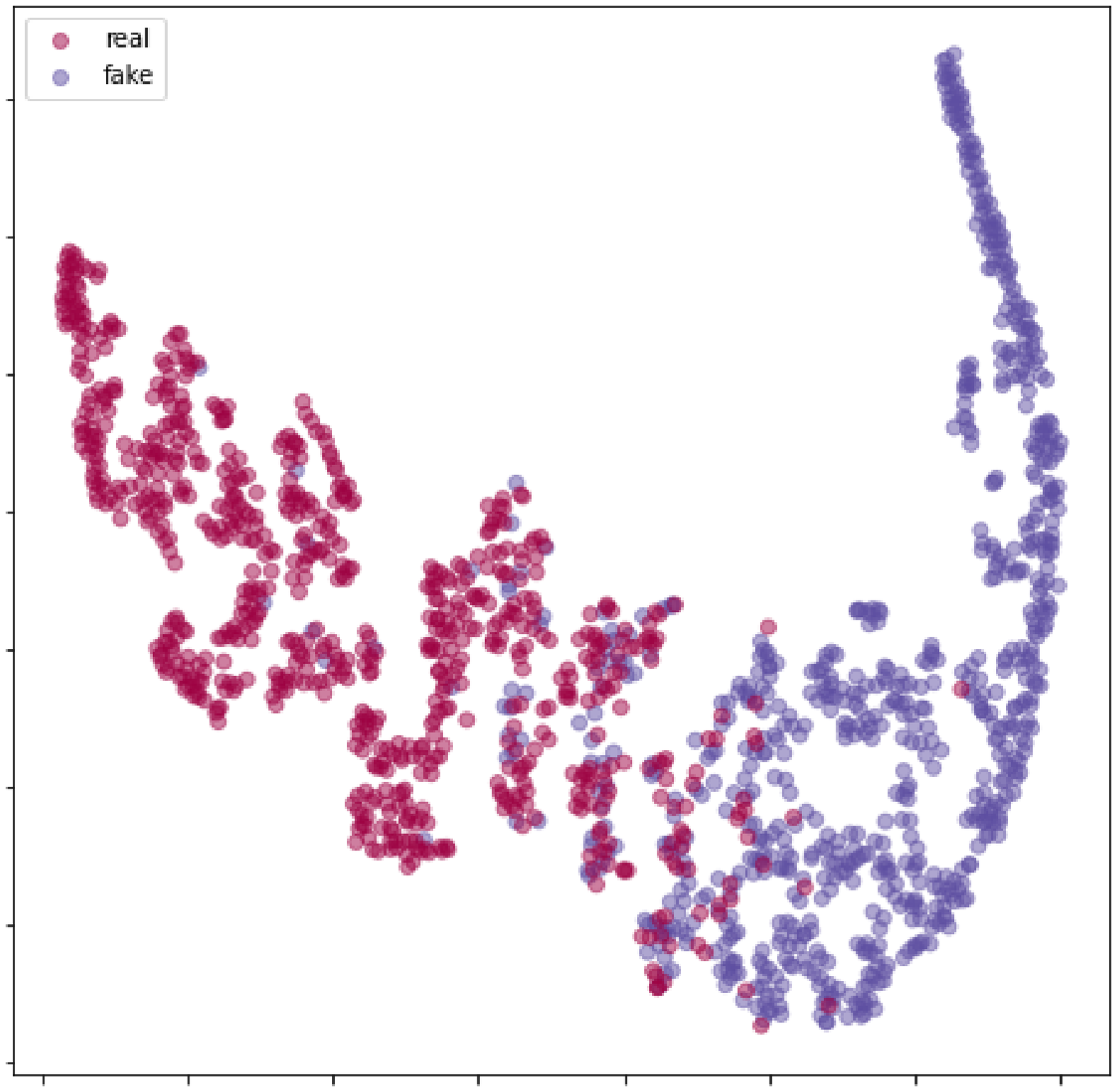}
		\label{fig:gat}
	}
	
	\subfloat[GAT2H]{
		\includegraphics[width=0.33\linewidth,height=15em]{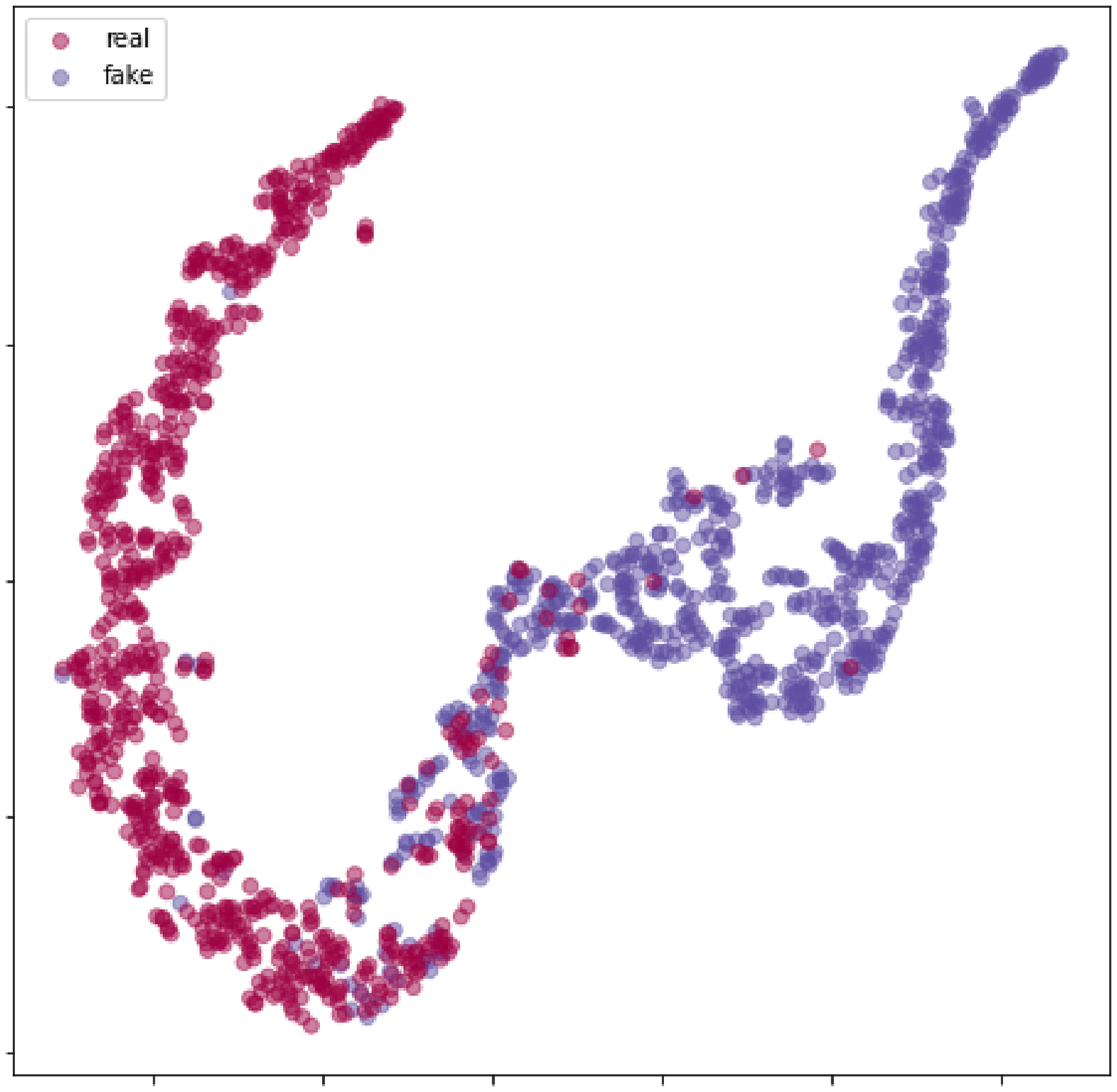}
		\label{fig:gat2h}
	}
	\subfloat[SemSeq4FD]{
		\includegraphics[width=0.33\linewidth,height=15em]{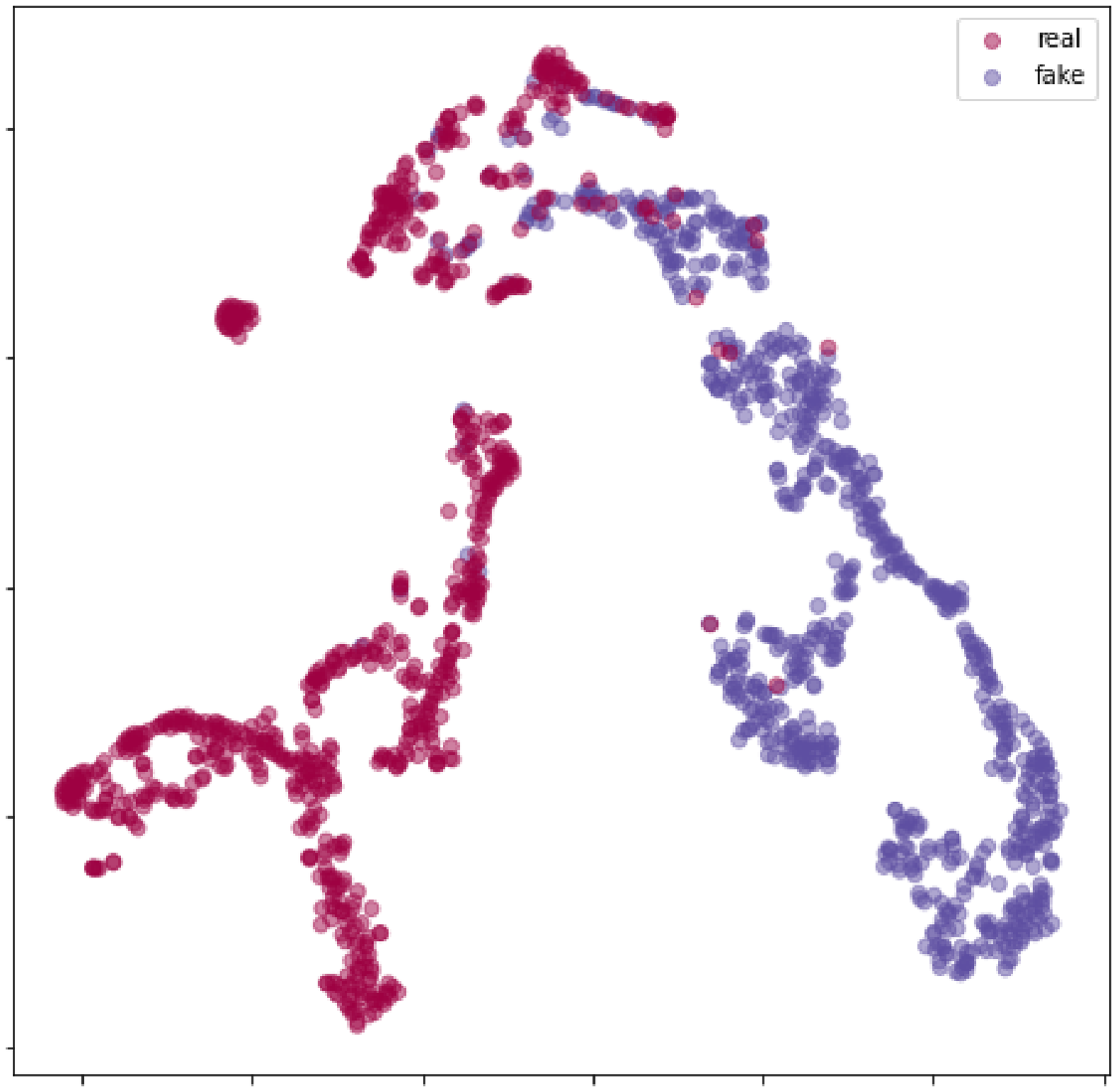}
		\label{fig:semseq4fd}
	}
	\subfloat[EDU4FD]{
		\includegraphics[width=0.33\linewidth,height=15em]{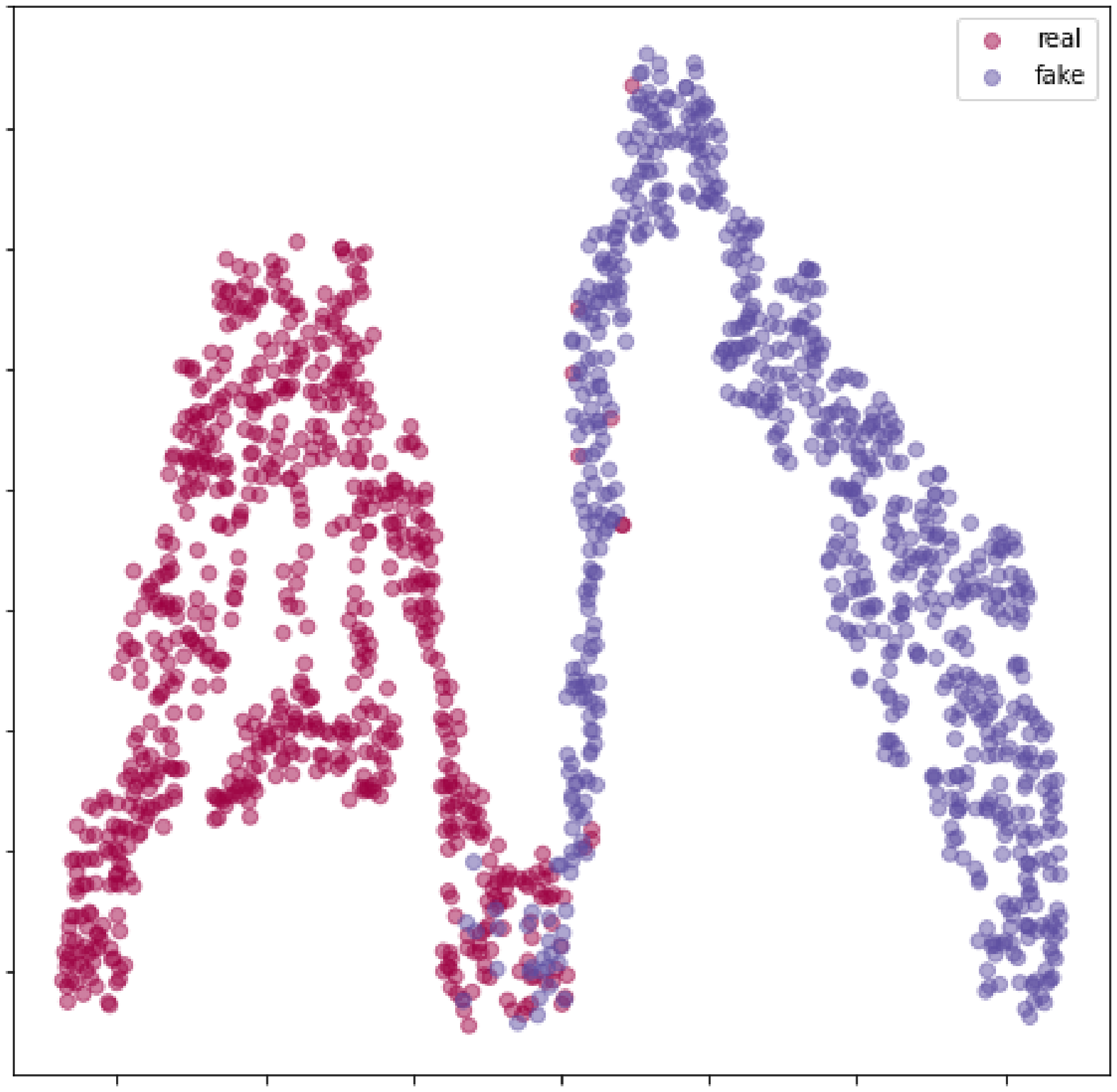}
		\label{fig:edu4fd}
	}
	\caption{The visualizations of text representation of EDU4FD and baselines.}
	\label{fig: tsne}
\end{figure*}
To better show the high-quality representation of EDU4FD over other methods, we took news texts from LUN-test and visualized text representations learned from EDU4FD and other baselines. Specifically, we first obtained the 100-dimensional text representations before the final Softmax layer, then used the t-Distributed Stochastic Neighbor Embedding (t-SNE) \citep{maaten2008visualizing} for visualizing. The 6 scatter plots are shown in Figure \ref{fig: tsne}, where red and blue dots correspond to real and fake text labels respectively. 
From the visualization results we can find that:
\begin{itemize}
\item Compared with the BERT model, the graph-based deep learning network models clearly separate texts as two clusters, demonstrating that using the structural information within text could help improve the effect of fake news detection.
\item Compared with the other five baseline models, the EDU4FD model proposed in this paper performs best and shows stronger semantic expression ability.
\end{itemize}
\section{Discussion}
Text-based detection methods can effectively detect fake news at an early stage. Previous text-based methods tend to focus on words or sentences level. However, isolated words lack coherence information, and long sentences are too coarse-grained and redundant to convey more specific information. The Elementary Discourse Unit (EDU) is the intermediate granularity between them and provides a better option to enhance the text representation. Meanwhile, our proposed model exploits EDU structures from different views to help represent text semantic. 

However, there are two limitations of our model that should be improved in the future. (1) The pre-processing steps of EDU segmentation and dependency graph construction are complicated and time-consumed. We still need to work on a simple and effective approach. (2) In addition to EDU, the rhetorical relations with specific functional meanings are also important for expressing text content, such as cause and temporal relations. Research work should be proposed to explore the semantics of rhetorical relations to further improve the text representation.

\section{Conclusion}
In this study, a novel multi-EDU-structure awareness fake news detection model named EDU4FD is proposed. It includes six major modules: EDU segmentation and dependency graph construction module, EDUs embedding module, sequence-based representation learning module, graph-based representation learning module, GRU-GA-based text representation module and inductive classification module. These components first segment a text into EDUs and construct the dependency graph, then obtain sequence-based and graph-based EDU representations, which are finally seamlessly integrated into an enhanced text representation for classification. Experimental results on four cross-source fake news datasets demonstrate that EDU4FD has excellent performance, indicating that using EDU and its multi-structures is significant for detecting fake news.

The various forms of fake news make automatic early detection more challenging. In the future, we plan to further leverage diverse types of data, and apply multimodal fusion method to identify fake news with forged images.

\section{Acknowledgements}
This work was supported by the National Natural Science Foundation of China (No: 61872260) and National key research and development program of China (No: 2021YFB3300503).
%
\bibliographystyle{model5-names}

\bibliography{references}  

\begin{thebibliography}{37}
\expandafter\ifx\csname natexlab\endcsname\relax\def\natexlab#1{#1}\fi
\providecommand{\url}[1]{\texttt{#1}}
\providecommand{\href}[2]{#2}
\providecommand{\path}[1]{#1}
\providecommand{\DOIprefix}{doi:}
\providecommand{\ArXivprefix}{arXiv:}
\providecommand{\URLprefix}{URL: }
\providecommand{\Pubmedprefix}{pmid:}
\providecommand{\doi}[1]{\href{http://dx.doi.org/#1}{\path{#1}}}
\providecommand{\Pubmed}[1]{\href{pmid:#1}{\path{#1}}}
\providecommand{\bibinfo}[2]{#2}
\ifx\xfnm\relax \def\xfnm[#1]{\unskip,\space#1}\fi
\bibitem[{{Ahn} \& {Jeong}(2019)}]{8864171}
\bibinfo{author}{{Ahn}, Y.}, \& \bibinfo{author}{{Jeong}, C.}
  (\bibinfo{year}{2019}).
\newblock \bibinfo{title}{Natural language contents evaluation system for
  detecting fake news using deep learning}.
\newblock In {\it \bibinfo{booktitle}{2019 16th International Joint Conference
  on Computer Science and Software Engineering (JCSSE)}\/} (pp.
  \bibinfo{pages}{289--292}).
\bibitem[{Cho et~al.(2014{\natexlab{a}})Cho, van Merri{\"e}nboer, Gulcehre,
  Bahdanau, Bougares, Schwenk \& Bengio}]{cho-etal-2014-learning}
\bibinfo{author}{Cho, K.}, \bibinfo{author}{van Merri{\"e}nboer, B.},
  \bibinfo{author}{Gulcehre, C.}, \bibinfo{author}{Bahdanau, D.},
  \bibinfo{author}{Bougares, F.}, \bibinfo{author}{Schwenk, H.}, \&
  \bibinfo{author}{Bengio, Y.} (\bibinfo{year}{2014}{\natexlab{a}}).
\newblock \bibinfo{title}{Learning phrase representations using {RNN}
  encoder{--}decoder for statistical machine translation}.
\newblock In {\it \bibinfo{booktitle}{EMNLP '14}\/} (pp.
  \bibinfo{pages}{1724--1734}).
\newblock \bibinfo{address}{Doha, Qatar}: \bibinfo{publisher}{Association for
  Computational Linguistics}.
\newblock \URLprefix \url{https://www.aclweb.org/anthology/D14-1179}.
  \DOIprefix\doi{10.3115/v1/D14-1179}.
\bibitem[{Cho et~al.(2014{\natexlab{b}})Cho, van Merri{\"e}nboer, Gulcehre,
  Bahdanau, Bougares, Schwenk \& Bengio}]{cho2014learning}
\bibinfo{author}{Cho, K.}, \bibinfo{author}{van Merri{\"e}nboer, B.},
  \bibinfo{author}{Gulcehre, C.}, \bibinfo{author}{Bahdanau, D.},
  \bibinfo{author}{Bougares, F.}, \bibinfo{author}{Schwenk, H.}, \&
  \bibinfo{author}{Bengio, Y.} (\bibinfo{year}{2014}{\natexlab{b}}).
\newblock \bibinfo{title}{Learning phrase representations using rnn
  encoder--decoder for statistical machine translation}.
\newblock In {\it \bibinfo{booktitle}{Proceedings of the 2014 Conference on
  Empirical Methods in Natural Language Processing (EMNLP)}\/} (pp.
  \bibinfo{pages}{1724--1734}).
\bibitem[{Devlin et~al.(2019)Devlin, Chang, Lee \&
  Toutanova}]{devlin-etal-2019-bert}
\bibinfo{author}{Devlin, J.}, \bibinfo{author}{Chang, M.-W.},
  \bibinfo{author}{Lee, K.}, \& \bibinfo{author}{Toutanova, K.}
  (\bibinfo{year}{2019}).
\newblock \bibinfo{title}{{BERT}: Pre-training of deep bidirectional
  transformers for language understanding}.
\newblock In {\it \bibinfo{booktitle}{NAACL '19}\/} (pp.
  \bibinfo{pages}{4171--4186}).
\bibitem[{Goldani et~al.(2021)Goldani, Safabakhsh \&
  Momtazi}]{GOLDANI2021102418}
\bibinfo{author}{Goldani, M.~H.}, \bibinfo{author}{Safabakhsh, R.}, \&
  \bibinfo{author}{Momtazi, S.} (\bibinfo{year}{2021}).
\newblock \bibinfo{title}{Convolutional neural network with margin loss for
  fake news detection}.
\newblock {\it \bibinfo{journal}{Information Processing \& Management}\/},
  {\it \bibinfo{volume}{58}\/}, \bibinfo{pages}{102418}. \URLprefix
  \url{https://www.sciencedirect.com/science/article/pii/S0306457320309134}.
  \DOIprefix\doi{https://doi.org/10.1016/j.ipm.2020.102418}.
\bibitem[{Horne \& Adali(2017)}]{horne2017just}
\bibinfo{author}{Horne, B.~D.}, \& \bibinfo{author}{Adali, S.}
  (\bibinfo{year}{2017}).
\newblock \bibinfo{title}{This just in: Fake news packs a lot in title, uses
  simpler, repetitive content in text body, more similar to satire than real
  news}.
\newblock In {\it \bibinfo{booktitle}{ICWSM '17}\/}.
\bibitem[{Ji \& Eisenstein(2014)}]{ji2014representation}
\bibinfo{author}{Ji, Y.}, \& \bibinfo{author}{Eisenstein, J.}
  (\bibinfo{year}{2014}).
\newblock \bibinfo{title}{Representation learning for text-level discourse
  parsing}.
\newblock In {\it \bibinfo{booktitle}{Proceedings of the 52nd annual meeting of
  the association for computational linguistics (volume 1: Long papers)}\/}
  (pp. \bibinfo{pages}{13--24}).
\bibitem[{Kim(2014)}]{kim-2014-convolutional}
\bibinfo{author}{Kim, Y.} (\bibinfo{year}{2014}).
\newblock \bibinfo{title}{Convolutional neural networks for sentence
  classification}.
\newblock In {\it \bibinfo{booktitle}{EMNLP '14}\/} (pp.
  \bibinfo{pages}{1746--1751}).
\bibitem[{Kipf \& Welling(2017)}]{DBLP:conf/iclr/KipfW17}
\bibinfo{author}{Kipf, T.~N.}, \& \bibinfo{author}{Welling, M.}
  (\bibinfo{year}{2017}).
\newblock \bibinfo{title}{Semi-supervised classification with graph
  convolutional networks}.
\newblock In {\it \bibinfo{booktitle}{{ICLR} 2017}\/}.
\bibitem[{Kleinbaum et~al.(2002)Kleinbaum, Dietz, Gail, Klein \&
  Klein}]{kleinbaum2002logistic}
\bibinfo{author}{Kleinbaum, D.~G.}, \bibinfo{author}{Dietz, K.},
  \bibinfo{author}{Gail, M.}, \bibinfo{author}{Klein, M.}, \&
  \bibinfo{author}{Klein, M.} (\bibinfo{year}{2002}).
\newblock {\it \bibinfo{title}{Logistic Regression}\/}.
\newblock \bibinfo{publisher}{Springer}.
\bibitem[{Li et~al.(2014)Li, Wang, Cao \& Li}]{li-etal-2014-text}
\bibinfo{author}{Li, S.}, \bibinfo{author}{Wang, L.}, \bibinfo{author}{Cao,
  Z.}, \& \bibinfo{author}{Li, W.} (\bibinfo{year}{2014}).
\newblock \bibinfo{title}{Text-level discourse dependency parsing}.
\newblock In {\it \bibinfo{booktitle}{ACL '14}\/} (pp.
  \bibinfo{pages}{25--35}).
\newblock \bibinfo{publisher}{Association for Computational Linguistics}.
\bibitem[{Luong et~al.(2015)Luong, Pham \& Manning}]{luong-etal-2015-effective}
\bibinfo{author}{Luong, T.}, \bibinfo{author}{Pham, H.}, \&
  \bibinfo{author}{Manning, C.~D.} (\bibinfo{year}{2015}).
\newblock \bibinfo{title}{Effective approaches to attention-based neural
  machine translation}.
\newblock In {\it \bibinfo{booktitle}{EMNLP '15}\/} (pp.
  \bibinfo{pages}{1412--1421}).
\newblock \bibinfo{address}{Lisbon, Portugal}.
\newblock \DOIprefix\doi{10.18653/v1/D15-1166}.
\bibitem[{Ma et~al.(2015)Ma, Gao, Wei, Lu \& Wong}]{10.1145/2806416.2806607}
\bibinfo{author}{Ma, J.}, \bibinfo{author}{Gao, W.}, \bibinfo{author}{Wei, Z.},
  \bibinfo{author}{Lu, Y.}, \& \bibinfo{author}{Wong, K.-F.}
  (\bibinfo{year}{2015}).
\newblock \bibinfo{title}{Detect rumors using time series of social context
  information on microblogging websites}.
\newblock In {\it \bibinfo{booktitle}{CIKM '15}\/} CIKM ’15 (p.
  \bibinfo{pages}{1751–1754}).
\newblock \bibinfo{address}{New York, NY, USA}: \bibinfo{publisher}{Association
  for Computing Machinery}.
\newblock \URLprefix \url{https://doi.org/10.1145/2806416.2806607}.
  \DOIprefix\doi{10.1145/2806416.2806607}.
\bibitem[{Maaten \& Hinton(2008)}]{maaten2008visualizing}
\bibinfo{author}{Maaten, v. d.~L.}, \& \bibinfo{author}{Hinton, G.}
  (\bibinfo{year}{2008}).
\newblock \bibinfo{title}{Visualizing data using t-sne}.
\newblock {\it \bibinfo{journal}{JOURNAL OF MACHINE LEARNING RESEARCH}\/},
  (pp. \bibinfo{pages}{2579--2605}).
\bibitem[{Mann \& Thompson(1988)}]{mann1988rhetorical}
\bibinfo{author}{Mann, W.~C.}, \& \bibinfo{author}{Thompson, S.~A.}
  (\bibinfo{year}{1988}).
\newblock \bibinfo{title}{Rhetorical structure theory: Toward a functional
  theory of text organization}.
\newblock {\it \bibinfo{journal}{Text-interdisciplinary Journal for the Study
  of Discourse}\/},  {\it \bibinfo{volume}{8}\/}, \bibinfo{pages}{243--281}.
\bibitem[{P{\'e}rez-Rosas et~al.(2018)P{\'e}rez-Rosas, Kleinberg, Lefevre \&
  Mihalcea}]{perezrosas2018automatic}
\bibinfo{author}{P{\'e}rez-Rosas, V.}, \bibinfo{author}{Kleinberg, B.},
  \bibinfo{author}{Lefevre, A.}, \& \bibinfo{author}{Mihalcea, R.}
  (\bibinfo{year}{2018}).
\newblock \bibinfo{title}{Automatic detection of fake news}.
\newblock In {\it \bibinfo{booktitle}{COLING '18}\/} (pp.
  \bibinfo{pages}{3391--3401}).
\bibitem[{Rashkin et~al.(2017)Rashkin, Choi, Jang, Volkova \&
  Choi}]{rashkin-etal-2017-truth}
\bibinfo{author}{Rashkin, H.}, \bibinfo{author}{Choi, E.},
  \bibinfo{author}{Jang, J.~Y.}, \bibinfo{author}{Volkova, S.}, \&
  \bibinfo{author}{Choi, Y.} (\bibinfo{year}{2017}).
\newblock \bibinfo{title}{Truth of varying shades: Analyzing language in fake
  news and political fact-checking}.
\newblock In {\it \bibinfo{booktitle}{EMNLP '17}\/} (pp.
  \bibinfo{pages}{2931--2937}).
\bibitem[{Rubin et~al.(2016)Rubin, Conroy, Chen \& Cornwell}]{rubin2016fake}
\bibinfo{author}{Rubin, V.~L.}, \bibinfo{author}{Conroy, N.},
  \bibinfo{author}{Chen, Y.}, \& \bibinfo{author}{Cornwell, S.}
  (\bibinfo{year}{2016}).
\newblock \bibinfo{title}{Fake news or truth? using satirical cues to detect
  potentially misleading news}.
\newblock In {\it \bibinfo{booktitle}{Proceedings of the second workshop on
  computational approaches to deception detection}\/} (pp.
  \bibinfo{pages}{7--17}).
\bibitem[{Rubin \& Lukoianova(2015)}]{10.1002/asi.23216}
\bibinfo{author}{Rubin, V.~L.}, \& \bibinfo{author}{Lukoianova, T.}
  (\bibinfo{year}{2015}).
\newblock \bibinfo{title}{Truth and deception at the rhetorical structure
  level}.
\newblock {\it \bibinfo{journal}{J. Assoc. Inf. Sci. Technol.}\/},  {\it
  \bibinfo{volume}{66}\/}, \bibinfo{pages}{905–917}. \URLprefix
  \url{https://doi.org/10.1002/asi.23216}. \DOIprefix\doi{10.1002/asi.23216}.
\bibitem[{Schlichtkrull et~al.(2018)Schlichtkrull, Kipf, Bloem, Van Den~Berg,
  Titov \& Welling}]{schlichtkrull2018modeling}
\bibinfo{author}{Schlichtkrull, M.}, \bibinfo{author}{Kipf, T.~N.},
  \bibinfo{author}{Bloem, P.}, \bibinfo{author}{Van Den~Berg, R.},
  \bibinfo{author}{Titov, I.}, \& \bibinfo{author}{Welling, M.}
  (\bibinfo{year}{2018}).
\newblock \bibinfo{title}{Modeling relational data with graph convolutional
  networks}.
\newblock In {\it \bibinfo{booktitle}{European Semantic Web Conference}\/} (pp.
  \bibinfo{pages}{593--607}).
\newblock \bibinfo{organization}{Springer}.
\bibitem[{Scholkopf \& Smola(2001)}]{10.5555/559923}
\bibinfo{author}{Scholkopf, B.}, \& \bibinfo{author}{Smola, A.~J.}
  (\bibinfo{year}{2001}).
\newblock {\it \bibinfo{title}{Learning with Kernels: Support Vector Machines,
  Regularization, Optimization, and Beyond}\/}.
\newblock \bibinfo{address}{Cambridge, MA, USA}: \bibinfo{publisher}{MIT
  Press}.
\bibitem[{Shu et~al.(2019)Shu, Cui, Wang, Lee \& Liu}]{shu2019defend}
\bibinfo{author}{Shu, K.}, \bibinfo{author}{Cui, L.}, \bibinfo{author}{Wang,
  S.}, \bibinfo{author}{Lee, D.}, \& \bibinfo{author}{Liu, H.}
  (\bibinfo{year}{2019}).
\newblock \bibinfo{title}{Defend: Explainable fake news detection}.
\newblock In {\it \bibinfo{booktitle}{Proceedings of the 25th ACM SIGKDD
  International Conference on Knowledge Discovery \& Data Mining}\/} (pp.
  \bibinfo{pages}{395--405}).
\bibitem[{Shu et~al.(2017)Shu, Wang \& Liu}]{shu2017exploiting}
\bibinfo{author}{Shu, K.}, \bibinfo{author}{Wang, S.}, \& \bibinfo{author}{Liu,
  H.} (\bibinfo{year}{2017}).
\newblock \bibinfo{title}{Exploiting tri-relationship for fake news detection}.
\newblock {\it \bibinfo{journal}{arXiv preprint arXiv:1712.07709}\/}, .
\bibitem[{Shu et~al.(2018)Shu, Wang \& Liu}]{shu2018understanding}
\bibinfo{author}{Shu, K.}, \bibinfo{author}{Wang, S.}, \& \bibinfo{author}{Liu,
  H.} (\bibinfo{year}{2018}).
\newblock \bibinfo{title}{Understanding user profiles on social media for fake
  news detection}.
\newblock In {\it \bibinfo{booktitle}{2018 IEEE Conference on Multimedia
  Information Processing and Retrieval (MIPR)}\/} (pp.
  \bibinfo{pages}{430--435}).
\newblock \bibinfo{organization}{IEEE}.
\bibitem[{Vaibhav et~al.(2019)Vaibhav, Mandyam \&
  Hovy}]{vaibhav-etal-2019-sentence}
\bibinfo{author}{Vaibhav, V.}, \bibinfo{author}{Mandyam, R.}, \&
  \bibinfo{author}{Hovy, E.} (\bibinfo{year}{2019}).
\newblock \bibinfo{title}{Do sentence interactions matter? leveraging sentence
  level representations for fake news classification}.
\newblock In {\it \bibinfo{booktitle}{Proceedings of the Thirteenth Workshop on
  Graph-Based Methods for Natural Language Processing (TextGraphs-13)}\/} (pp.
  \bibinfo{pages}{134--139}).
\bibitem[{Veli{\v{c}}kovi{\'c} et~al.(2018)Veli{\v{c}}kovi{\'c}, Cucurull,
  Casanova, Romero, Li{\`o} \& Bengio}]{velivckovic2018graph}
\bibinfo{author}{Veli{\v{c}}kovi{\'c}, P.}, \bibinfo{author}{Cucurull, G.},
  \bibinfo{author}{Casanova, A.}, \bibinfo{author}{Romero, A.},
  \bibinfo{author}{Li{\`o}, P.}, \& \bibinfo{author}{Bengio, Y.}
  (\bibinfo{year}{2018}).
\newblock \bibinfo{title}{Graph attention networks}.
\newblock In {\it \bibinfo{booktitle}{International Conference on Learning
  Representations}\/}.
\bibitem[{Volkova et~al.(2017)Volkova, Shaffer, Jang \&
  Hodas}]{volkova-etal-2017-separating}
\bibinfo{author}{Volkova, S.}, \bibinfo{author}{Shaffer, K.},
  \bibinfo{author}{Jang, J.~Y.}, \& \bibinfo{author}{Hodas, N.}
  (\bibinfo{year}{2017}).
\newblock \bibinfo{title}{Separating facts from fiction: Linguistic models to
  classify suspicious and trusted news posts on twitter}.
\newblock In {\it \bibinfo{booktitle}{ACL '17}\/} (pp.
  \bibinfo{pages}{647--653}).
\newblock \bibinfo{address}{Vancouver, Canada}.
\bibitem[{Wang(2022)}]{LiWang_14}
\bibinfo{author}{Wang, L.} (\bibinfo{year}{2022}).
\newblock \bibinfo{title}{Development and prospect of false information
  detection on social medias}.
\newblock {\it \bibinfo{journal}{Journal of Taiyuan University of
  Technology}\/},  (pp. \bibinfo{pages}{1--14}).
\bibitem[{Wang(2017)}]{wang-2017-liar}
\bibinfo{author}{Wang, W.~Y.} (\bibinfo{year}{2017}).
\newblock \bibinfo{title}{{``}liar, liar pants on fire{''}: A new benchmark
  dataset for fake news detection}.
\newblock In {\it \bibinfo{booktitle}{ACL '17}\/} (pp.
  \bibinfo{pages}{422--426}).
\bibitem[{Wang et~al.(2021)Wang, Wang, Yang \& Lian}]{WANG2021114090}
\bibinfo{author}{Wang, Y.}, \bibinfo{author}{Wang, L.}, \bibinfo{author}{Yang,
  Y.}, \& \bibinfo{author}{Lian, T.} (\bibinfo{year}{2021}).
\newblock \bibinfo{title}{Semseq4fd: Integrating global semantic relationship
  and local sequential order to enhance text rep-resentation for fake news
  detection}.
\newblock {\it \bibinfo{journal}{Expert Systems with Applications}\/},  {\it
  \bibinfo{volume}{166}\/}, \bibinfo{pages}{114090}.
\bibitem[{Xue et~al.(2021)Xue, Wang, Yang \& Lian}]{Lianbiao_3540}
\bibinfo{author}{Xue, H.}, \bibinfo{author}{Wang, L.}, \bibinfo{author}{Yang,
  Y.}, \& \bibinfo{author}{Lian, b.} (\bibinfo{year}{2021}).
\newblock \bibinfo{title}{Rumor detection model based on user propagation
  network and message content}.
\newblock {\it \bibinfo{journal}{Journal of Computer Applications}\/},  {\it
  \bibinfo{volume}{41}\/}, \bibinfo{pages}{3540--3545}.
\bibitem[{Yang et~al.(2021)Yang, Wang \& Wang}]{20212910664304}
\bibinfo{author}{Yang, Y.}, \bibinfo{author}{Wang, L.}, \&
  \bibinfo{author}{Wang, Y.} (\bibinfo{year}{2021}).
\newblock \bibinfo{title}{Rumor detection based on source information and
  gating graph neural network}.
\newblock {\it \bibinfo{journal}{Computer Research and Development}\/},  {\it
  \bibinfo{volume}{58}\/}, \bibinfo{pages}{1412 -- 1424}. \URLprefix
  \url{http://dx.doi.org/10.7544/issn1000-1239.2021.20200801}.
\bibitem[{Yang et~al.(2022)Yang, Wang, Wang \& Meng}]{YANG2022116071}
\bibinfo{author}{Yang, Y.}, \bibinfo{author}{Wang, Y.}, \bibinfo{author}{Wang,
  L.}, \& \bibinfo{author}{Meng, J.} (\bibinfo{year}{2022}).
\newblock \bibinfo{title}{Postcom2dr: Utilizing information from post and
  comments to detect rumors}.
\newblock {\it \bibinfo{journal}{Expert Systems with Applications}\/},  {\it
  \bibinfo{volume}{189}\/}, \bibinfo{pages}{116071}. \URLprefix
  \url{https://www.sciencedirect.com/science/article/pii/S095741742101410X}.
  \DOIprefix\doi{https://doi.org/10.1016/j.eswa.2021.116071}.
\bibitem[{Yao et~al.(2019)Yao, Mao \& Luo}]{yao2019graph}
\bibinfo{author}{Yao, L.}, \bibinfo{author}{Mao, C.}, \& \bibinfo{author}{Luo,
  Y.} (\bibinfo{year}{2019}).
\newblock \bibinfo{title}{Graph convolutional networks for text
  classification}.
\newblock In {\it \bibinfo{booktitle}{"AAAI '19"}\/} (pp.
  \bibinfo{pages}{7370--7377}).
\newblock volume~\bibinfo{volume}{33}.
\bibitem[{Yu et~al.(2017)Yu, Liu, Wu, Wang \& Tan}]{ijcai2017-545}
\bibinfo{author}{Yu, F.}, \bibinfo{author}{Liu, Q.}, \bibinfo{author}{Wu, S.},
  \bibinfo{author}{Wang, L.}, \& \bibinfo{author}{Tan, T.}
  (\bibinfo{year}{2017}).
\newblock \bibinfo{title}{A convolutional approach for misinformation
  identification}.
\newblock In {\it \bibinfo{booktitle}{Proceedings of the Twenty-Sixth
  International Joint Conference on Artificial Intelligence, {IJCAI-17}}\/}
  (pp. \bibinfo{pages}{3901--3907}).
\newblock \URLprefix \url{https://doi.org/10.24963/ijcai.2017/545}.
  \DOIprefix\doi{10.24963/ijcai.2017/545}.
\bibitem[{Zhang et~al.(2020)Zhang, Yu, Cui, Wu, Wen \&
  Wang}]{zhang-etal-2020-every}
\bibinfo{author}{Zhang, Y.}, \bibinfo{author}{Yu, X.}, \bibinfo{author}{Cui,
  Z.}, \bibinfo{author}{Wu, S.}, \bibinfo{author}{Wen, Z.}, \&
  \bibinfo{author}{Wang, L.} (\bibinfo{year}{2020}).
\newblock \bibinfo{title}{Every document owns its structure: Inductive text
  classification via graph neural networks}.
\newblock In {\it \bibinfo{booktitle}{ACL}\/} (pp. \bibinfo{pages}{334--339}).
\bibitem[{Zhou et~al.(2020)Zhou, Jain, Phoha \& Zafarani}]{zhou2020fake}
\bibinfo{author}{Zhou, X.}, \bibinfo{author}{Jain, A.}, \bibinfo{author}{Phoha,
  V.~V.}, \& \bibinfo{author}{Zafarani, R.} (\bibinfo{year}{2020}).
\newblock \bibinfo{title}{Fake news early detection: A theory-driven model}.
\newblock {\it \bibinfo{journal}{Digital Threats: Research and Practice}\/},
  {\it \bibinfo{volume}{1}\/}, \bibinfo{pages}{1--25}.

\end{thebibliography}
\section*{Appendix} \label{appendix}
There are 19 kinds of rhetorical relations mentioned in \citep{li-etal-2014-text} have been used in this paper. We summarize the relations' statistic in Table \ref{tab:relation_number1} and Table \ref{tab:relation_number2}, indicating the frequency of each relation in the training and test corpus.
\begin{table*}[h]
	\centering
	\caption{Relation Distribution of the LUN-train, LUN-test and SLN datasets}
	\label{tab:relation_number1}
	\renewcommand\arraystretch{0.9}
	\begin{tabular}{l|l|l|l|l|l|l}
		\hline
		\multicolumn{1}{c|}{relation} & \multicolumn{2}{|c|}{LUN-train}                       & \multicolumn{2}{|c|}{LUN-test}                        & \multicolumn{2}{|c}{SLN}                             \\ \hline
		& \multicolumn{1}{|c|}{Real} & \multicolumn{1}{|c|}{Fake} & \multicolumn{1}{|c|}{Real} & \multicolumn{1}{|c|}{Fake} & \multicolumn{1}{|c|}{Real} & \multicolumn{1}{|c}{Fake} \\ \hline
		Topic-comment                & 0.006                    & 0.011                    & 0.006                    & 0.010                    & 0.006                    & 0.011                    \\
		Topic-change                 & 0.002                    & 0.001                    & 0.001                    & 0.001                    & 0.001                    & 0.001                    \\
		Textual                      & 0.039                    & 0.023                    & 0.034                    & 0.020                    & 0.032                    & 0.020                    \\
		Temporal                     & 0.025                    & 0.020                    & 0.025                    & 0.022                    & 0.027                    & 0.021                    \\
		Summary                      & 0.007                    & 0.004                    & 0.007                    & 0.006                    & 0.007                    & 0.005                    \\
		Same-unit                    & 0.012                    & 0.017                    & 0.011                    & 0.014                    & 0.010                    & 0.015                    \\
		Manner-means                 & 0.006                    & 0.008                    & 0.007                    & 0.008                    & 0.007                    & 0.007                    \\
		Joint                        & 0.021                    & 0.020                    & 0.021                    & 0.018                    & 0.021                    & 0.019                    \\
		Explanation                  & 0.010                    & 0.012                    & 0.010                    & 0.014                    & 0.009                    & 0.012                    \\
		Evaluation                   & 0.192                    & 0.157                    & 0.188                    & 0.137                    & 0.195                    & 0.143                    \\
		Root                         & 0.016                    & 0.016                    & 0.015                    & 0.017                    & 0.014                    & 0.016                    \\
		Enablement                   & 0.106                    & 0.120                    & 0.102                    & 0.135                    & 0.104                    & 0.134                    \\
		Elaboration                  & 0.294                    & 0.218                    & 0.306                    & 0.240                    & 0.301                    & 0.235                    \\
		Contrast                     & 0.074                    & 0.116                    & 0.074                    & 0.118                    & 0.074                    & 0.117                    \\
		Condition                    & 0.008                    & 0.014                    & 0.008                    & 0.013                    & 0.008                    & 0.015                    \\
		Comparison                   & 0.011                    & 0.011                    & 0.010                    & 0.011                    & 0.010                    & 0.011                    \\
		Cause                        & 0.012                    & 0.012                    & 0.013                    & 0.012                    & 0.012                    & 0.011                    \\
		Background                   & 0.014                    & 0.019                    & 0.014                    & 0.015                    & 0.013                    & 0.016                    \\
		Attribution                  & 0.145                    & 0.201                    & 0.146                    & 0.190                    & 0.147                    & 0.190  
		\\
		\hline                 
	\end{tabular}
\end{table*}

\begin{table*}[t]
	\centering
	\caption{Relation Distribution of the Kaggle, BuzzFeed and PolitiFact datasets}
	\label{tab:relation_number2}
	\renewcommand\arraystretch{0.9}
	\begin{tabular}{l|l|l|l|l|l|l}
		\hline
		\multicolumn{1}{c|}{relation} & \multicolumn{2}{c|}{Kaggle}                          & \multicolumn{2}{c|}{BuzzFeed}                        & \multicolumn{2}{c}{PolitiFact}                      \\ \hline
		\multicolumn{1}{c|}{}         & \multicolumn{1}{c|}{Real} & \multicolumn{1}{c|}{Fake} & \multicolumn{1}{c|}{Real} & \multicolumn{1}{c|}{Fake} & \multicolumn{1}{c|}{Real} & \multicolumn{1}{c}{Fake} \\ \hline
		Topic-comment                & 0.005                    & 0.004                    & 0.008                    & 0.006                    & 0.008                    & 0.009                    \\
		Topic-change                 & 0.002                    & 0.003                    & 0.002                    & 0.001                    & 0.001                    & 0.002                    \\
		Textual                      & 0.056                    & 0.073                    & 0.023                    & 0.017                    & 0.028                    & 0.025                    \\
		Temporal                     & 0.034                    & 0.048                    & 0.018                    & 0.020                    & 0.031                    & 0.024                    \\
		Summary                      & 0.008                    & 0.014                    & 0.007                    & 0.006                    & 0.010                    & 0.004                    \\
		Same-unit                    & 0.004                    & 0.004                    & 0.005                    & 0.010                    & 0.007                    & 0.007                    \\
		Manner-means                 & 0.006                    & 0.008                    & 0.008                    & 0.007                    & 0.004                    & 0.008                    \\
		Joint                        & 0.016                    & 0.017                    & 0.021                    & 0.029                    & 0.014                    & 0.027                    \\
		Explanation                  & 0.009                    & 0.008                    & 0.016                    & 0.011                    & 0.013                    & 0.014                    \\
		Evaluation                   & 0.210                    & 0.189                    & 0.229                    & 0.190                    & 0.192                    & 0.166                    \\
		Root                         & 0.015                    & 0.009                    & 0.014                    & 0.008                    & 0.009                    & 0.012                    \\
		Enablement                   & 0.106                    & 0.120                    & 0.108                    & 0.117                    & 0.113                    & 0.122                    \\
		Elaboration                  & 0.299                    & 0.305                    & 0.259                    & 0.287                    & 0.315                    & 0.299                    \\
		Contrast                     & 0.060                    & 0.068                    & 0.086                    & 0.113                    & 0.079                    & 0.088                    \\
		Condition                    & 0.006                    & 0.006                    & 0.008                    & 0.011                    & 0.009                    & 0.015                    \\
		Comparison                   & 0.007                    & 0.011                    & 0.004                    & 0.008                    & 0.006                    & 0.009                    \\
		Cause                        & 0.009                    & 0.011                    & 0.007                    & 0.006                    & 0.009                    & 0.011                    \\
		Background                   & 0.009                    & 0.008                    & 0.010                    & 0.017                    & 0.012                    & 0.016                    \\
		Attribution                  & 0.137                    & 0.094                    & 0.165                    & 0.135                    & 0.138                    & 0.141            \\
		\hline       
	\end{tabular}
\end{table*}

\end{document}